\documentclass[final,3p,times]{elsarticle}

\usepackage{amsmath}
\usepackage{makecell}
\usepackage{graphicx}
\usepackage{tabularx}
\usepackage{amssymb}
\usepackage{subcaption}
\usepackage{threeparttable}
\usepackage{amsmath,lipsum}
\usepackage{cuted}
\usepackage{natbib}
\setcitestyle{square, comma, numbers,sort&compress}
\usepackage{multirow}
\usepackage{color}
\usepackage{float}
\usepackage{stfloats}
\usepackage{booktabs}

\newtheorem{assumption}{Assumption}
\newtheorem{example}{Example}
\newtheorem{property}{Property}
\newtheorem{proposition}{Proposition}
\usepackage[linesnumbered,ruled]{algorithm2e}
\usepackage{amsmath}
\journal{Elsevier}
\begin{document}
\begin{frontmatter}
\title{Exploring Over-stationarization in Deep Learning-based Bus/Tram Arrival Time Prediction: Analysis and Non-stationary Effect Recovery}

\author[label1]{Zirui Li}
\author[label2]{Bin Yang}
\author[label1]{Meng Wang\corref{cor1}}
\affiliation[label1]{organization={Chair of Traffic Process Automation, "Friedrich List" Faculty of Transport and Traffic Sciences, Technische Universität Dresden},
            addressline={Hettnerstraße 3},
            city={Dresden},
            postcode={01069},
            state={},
            country={Germany}}

\affiliation[label2]{organization={School of Data Science \& Engineering, East China Normal University},
            addressline={},
            city={Shanghai},
            postcode={200050},
            state={},
            country={China}}

\cortext[cor1]{Corresponding author: Meng Wang (Email address: meng.wang@tu-dresden.de)}

\begin{abstract}
Arrival time prediction (ATP) of public transport vehicles is essential in improving passenger experience and supporting traffic management. Deep learning has demonstrated outstanding performance in ATP abodel non-line and temporal dynamics. In the multi-step ATP, non-stationary data will degrade the model performance due to the variation in variables’ joint distribution along the temporal direction. Previous studies mainly applied normalization to eliminate the non-stationarity in time series, thereby achieving better predictability. However, the normalization may obscure useful characteristics inherent in non-stationarity, which is known as the over-stationarization. In this work, to trade off predictability and non-stationarity, a new approach for multi-step ATP, named non-stationary ATP ( NSATP), is proposed. The method consists of two stages: series stationarization and non-stationarity effect recovery. The first stage aims at improving the predictability. As for the latter, NSATP extends a state-of-the-art method from one-dimensional to two dimensional based models to capture the hidden periodicity in time series and designs a compensation module of over-stationarization by learning scaling and shifting factors from raw data. 125 days' public transport operational data of Dresden is collected for validation. Experimental results show that compared to baseline methods, the proposed NSATP can reduce RMSE, MAE, and MAPE by 2.37\%, 1.22\%, and 2.26\% for trams and by 1.72\%, 0.60\%, and 1.17\% for buses, respectively.
\end{abstract}


\begin{keyword}
Tram/bus arrival time\sep non-stationary effect\sep time series forecasting\sep series stationarization.
\end{keyword}
\end{frontmatter}

\section{Introduction}
Public transport (PT) plays a crucial role in alleviating urban traffic congestion and improving sustainability in big cities ~\cite{hensher2007sustainable,buehler2011making}. Public transport vehicles, such as buses and trams, often share infrastructure resources with private vehicles. Complex road conditions (e.g., congestion, traffic signals) and the uncertainty of passenger origin-destination flows can deviate PT vehicles from schedules and impair the reliability of public transport~\cite{soza2019underlying,amberg2019robust}. Accurate arrival time prediction (ATP) can improve passenger experiences and provide valuable information for reasonable travel choices~\cite{liu2025eco,lam2001value,lakhan2024multi}, and is an important component for multimodal traffic management systems.

\begin{figure}[t!]
  \centering
  \begin{subfigure}[b]{0.24\textwidth}
    \includegraphics[width=1\textwidth]{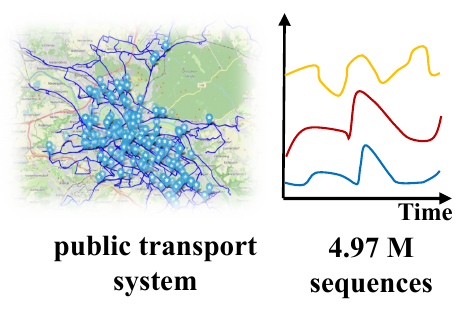}
    \caption{City-wide large-scale\\ data collection.}
    \label{introduction_fig_1}
  \end{subfigure}
  \begin{subfigure}[b]{0.24\textwidth}
    \includegraphics[width=1\textwidth]{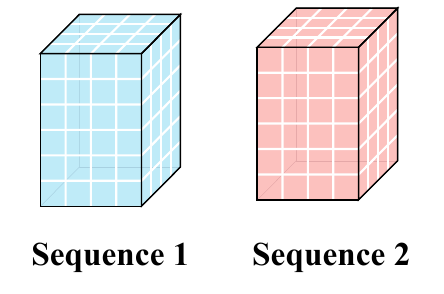}
    \caption{Two different raw sequences extracted from the dataset.}
    \label{introduction_fig_2}
  \end{subfigure}
  \begin{subfigure}[b]{0.24\textwidth}
    \includegraphics[width=1\textwidth]{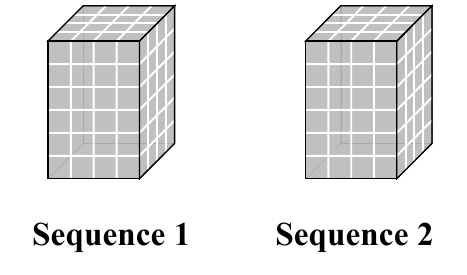}
    \caption{Two sequences after series stationarization are similar and indistinguishable.}
    \label{introduction_fig_3}
  \end{subfigure}
    \begin{subfigure}[b]{0.24\textwidth}
    \includegraphics[width=0.9\textwidth]{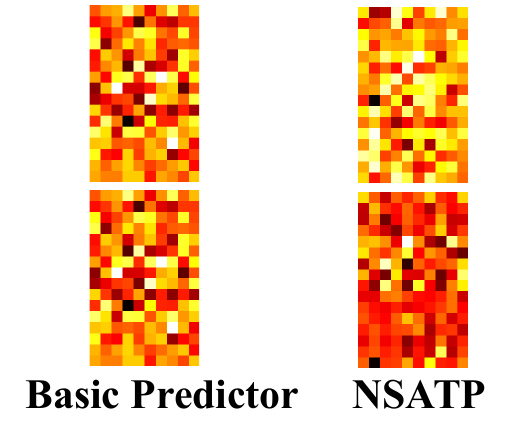}
    \caption{Features extracted by the basic predictor and the proposed NSATP.}
    \label{introduction_fig_4}
  \end{subfigure}
  \caption{The visualized illustration of over-stationarization. (a) presents the data collection from the central public transport management system. (b) is the showcase of two different raw sequences. (c) demonstrates that two normalized sequences are indistinguishable after the series stationarization. (d) illustrates that two similar sequences after the series stationarization cannot be recognized by the basic deep learning-based arrival time predictor, while the proposed NSATP can capture and recover the non-stationary effect.}
  \label{introduction_fig}
\end{figure}

Numerous approaches for ATP have been developed. Overall, these methods can be divided into three categories: state estimation-based, statistical learning-based, and deep learning-based methods.
The state estimation-based methods predict the future states of a dynamic system based on noisy measurements of its current and past states, which requires the existence of a direct mathematical connection between multiple variables~\cite{welch1995introduction,heckerman2008tutorial}. In~\cite{achar2019bus}, a novel Kalman filter (KF) was developed to learn the spatial and temporal correlations of public transport explicitly. It captured spatial patterns with a linear state-space function and updated temporal states based on historical data. To promote the predictability of ATP, a Bayesian hierarchical approach for constructing bus dwell time duration distributions from historical data was proposed in~\cite{isukapati2020hierarchical}. It relied on minimal data and provided a time-varying confidence for decision-making. In~\cite{chen2023probabilistic}, a probabilistic forecasting approach of bus travel time with a Bayesian Gaussian mixture model was presented, which utilized the adjacent information in the Bayesian update. Furthermore, a conditional travel time prediction based on Bayesian Markov regime-switching vector autoregression was proposed by considering the dynamic variation of passenger occupancy~\cite{chen2024conditional}. State estimation-based methods are highly time-efficient and require minimal data. However, these methods necessitate a clear physical relationship among various factors. Implicit influence and joint distributions among factors are hard to encode within state estimation algorithms.

Statistical learning-based methods, built on probability theory, aim to establish the mapping from inputs to outputs by analyzing data distributions~\cite{vapnik1999overview}. Typically, these relationships are nonlinear and involve multi-variable joint distributions. In~\cite{yu2011bus}, several statistical machine learning algorithms were adopted in the bus ATP at the same stop but with different routes, which included support vector machine (SVM), \textit{k} nearest neighbors algorithm (\textit{k}-NN) and linear regression (LR). In~\cite{ma2019bus}, a novel segment-based approach to automatically divide bus routes into dwelling and traveling segments was proposed. Then, two statistical learning models were applied to predict them separately by incorporating different impact factors. In~\cite{kumar2018hybrid}, a hybrid model for bus travel time estimation was presented. It identified significant inputs and used them in a hybrid model that combined the exponential smoothing technique with a recursive estimation scheme based on KF. Statistical learning-based methods are superior in their simple structure and high efficiency in training. However, most of these methods focus on single-step prediction, which cannot model the temporal variations of arrival time in sequential stops. Fortunately, it can be addressed by deep learning-based methods.

Several deep learning-based methods have been developed and applied in the ATP of public transport. In~\cite{he2020learning}, a Traffic Pattern centric Segment Coalescing Framework (TP-SCF), which learned disparate patterns of traffic conditions across different bus line segments, was proposed. A long-short term memory (LSTM) was applied for bus travel time prediction in each pattern. In~\cite{Ratneswaran}, a multi-model stacked ensemble learning strategy by leveraging the best-performing models was proposed. It could generate accurate predictions in dwell and running (travel) time by capturing heterogeneous traffic conditions~\cite{liu2025multi,el2011data}. In~\cite{liu2023understanding}, the statistical characteristics of travel time was analyzed based on historic AVL (Automatic Vehicle Location) data and developed a Kalman filter long-short term memory (KF-LSTM) to estimate bus travel time. Experimental results showed that KF-LSTM outperformed the ensemble learning strategy. Different from using multi-source heterogeneous information, in~\cite{li2023sequence}, a novel sequence and network embedding model for bus ATP using GPS (Global Positioning System) trajectory only was proposed, which provided real-time online prediction and monitoring for the public transport system. Most of the methods in bus/tram ATP focused on modeling the relationship in the temporal dimension. These approaches focused on capturing the pair-wise temporal dependencies among time points, which is termed one-dimensional based method. But it is hard to find out reliable dependencies directly from scattered time points, since the temporal dependencies  can be obscured deeply in intricate temporal patterns (e.g., rising, falling, and fluctuation)~\cite{hu2025jtfnet}. To cope with the issue above, \textit{ArrivalNet} was presented in~\cite{liarrivalnet}, which was a two-dimensional temporal variation-based multi-step ATP for buses and trams. It decomposed the one-dimensional temporal sequence into intra-periodic and inter-periodic variations~\cite{wu2022timesnet}. 

Among deep learning-based ATP methods, \textit{ArrivalNet} demonstrated superior performance in comparative experiments over methods like temporal convolutional network (TCN) and Transformer~\cite{bai2018empirical,vaswani2017attention}. Because of the stochastic passenger demand and the uncertainty of traffic conditions, feature values in public transport data may fluctuate markedly over time. This variability along the temporal dimension is known as non-stationarity, which will degrade the performance of deep learning models. In time series analysis, the feature-wise normalization along the temporal dimension is a widely adopted solution, commonly termed as series stationarization~\cite{liarrivalnet}. In other words, the process can be formulated as $\mathbf{X}^{\star} = \pmb{\alpha}\mathbf{X} + \pmb{\beta}$, where $\pmb{\alpha}$ and $\pmb{\beta}$ are scaling and shifting parameters in normalization. $\mathbf{X}$ and $\mathbf{X}^{\star}$ are original and standardized time series, respectively. 

However, a significant issue may arise in this process: two different time series, equipped with different means and variances in series stationarization, may be transformed into the same or similar sequences. The intrinsic non-stationary characteristics of the original sequences are removed, which renders the predictor unable to distinguish between two standardized time series. The phenomenon is known as \textbf{over-stationarization}~\cite{liu2022non}. As illustrated in Fig.~\ref{introduction_fig} the normalized sequences follow a more similar distribution compared to the original ones (Fig.~\ref{introduction_fig_4} left). Moreover, the model tends to produce overly stable and indistinguishable outputs, as it cannot capture the characteristics of non-stationarity.

To address this issue, in~\cite{kim2021reversible}, a two-stage instance normalization strategy was proposed, which designed learnable affine parameters for each input and restored the corresponding output. The analysis in~\cite{liu2022non} found that the model can still perform well even when the parameters are not learnable. Furthermore, an embedded de-stationary attention structure within the Transformer was presented, which recovered the specific temporal dependencies in non-stationary sequences in the calculation of the attention mechanism. The above strategy is termed NS Transformer. But it only focused on mitigating over-stationarization for one-dimensional modelling methods, e.g., Vanilla Transformer. We experimentally find that NS transformer is unsuitable for multi-step ATP, leading to gradient collapse and difficulty in convergence. This shortcoming may stem from the fact that one-dimensional modeling approaches cannot adequately capture the latent periodicity embedded in the time series of public transport. 

In this paper, to tackle the issue of over-stationarization in ATP, we analyze the influence of the non-stationarity in two-dimensional temporal variation-based time series predictor, which includes two kinds of mainstream variants in feature extraction: convolutional neural networks (CNN) and Swin Transformer~\cite{szegedy2015going,liu2021swin}. Based on the analysis, a non-stationary effect recovery approach is designed and embedded in DNN-based ATP, which is termed \textbf{N}on-\textbf{s}tationary \textbf{A}rrival \textbf{T}ime \textbf{P}redictor (NSATP). It consists of two steps: series stationarization and non-stationarity effect recovery. The first one is employed to improve model capability overall. The latter aims to capture the temporal dependencies of non-stationarity and recover them to avoid over-stationarization. NSATP overcomes the dilemma between time series predictability from stationarization and model capability in non-stationarity. To apply it in representative DNN architectures, two over-stationarization compensation approaches are designed to recover the non-stationarity of non-stationary series, which are formulated in CNN-based and transformer-based architectures, respectively. The proposed model is validated on city-wide public transport operational data.

The remainder of this paper is organized as follows. The problem formulation of ATP and the over-stationarization issue are detailed in Section~\ref{problem_formulation}. Section~\ref{ns_cnn} and Section~\ref{ns_attention} present two approaches for recovering non-stationarity characteristics. The experiments are shown in Section~\ref{experiments}, which includes the dataset description, experimental results and discussion. Finally, the conclusion and future work are summarized in Section~\ref{conclusion}.

\section{Problem formulation}\label{problem_formulation}
This work focuses on the non-stationary effect in the multi-step ATP of buses and trams. In this section, the formulation of ATP is first detailed. Then, series stationarization and the over-stationarization it introduces are analyzed. Finally, the core concept of the proposed NSATP is briefly presented.

\subsection{The formulation of multi-step bus/tram ATP}
Due to uncertainties stemming from road congestion, traffic signals, and passenger flow, the arrival time of buses and trams does not strictly adhere to the schedule. This leads to a dynamic delay at each stop, which can propagate to the next stop. Note that the delay can be either positive or negative. A negative delay means early arrival at the stop. For any given stop $i$, the multi-step ATP is described as follows:
\begin{equation}\label{prediction_eq}
\hat{\mathbf{T}}^{\text{d}}_{i+1:i+N_{\text{f}}} = f_{\pmb{\theta}}(\mathbf{F}^{\text{temporal}}_{i-N_{\text{p}}+1:i},\mathbf{F}^{\text{context}})
\end{equation}
\begin{equation}\label{multi_step}
    \hat{\mathbf{T}}^{\text{a}}_{i+1:i+N_{\text{f}}} = \hat{\mathbf{T}}^{\text{d}}_{i+1:i+N_{\text{f}}}+\mathbf{T}^{\text{s}}_{i+1:i+N_{\text{f}}}
\end{equation}
where $N_{\text{p}}$ and $N_{\text{f}}$ are the number of past and future stops, respectively. $\mathbf{F}^{\text{temporal}}_{i-N_{\text{p}}+1:i}$ is the temporal feature in the past $N_{\text{p}}$ stops. $\mathbf{F}^{\text{context}}$ is the contextual feature. $\mathbf{T}^{\text{s}}_{i+1:i+N_{\text{f}}}$, $\hat{\mathbf{T}}^{\text{d}}_{i+1:i+N_{\text{f}}}$ and $\hat{\mathbf{T}}^{\text{a}}_{i+1:i+N_{\text{f}}}$ are scheduled arrival times, predicted delays and predicted arrival times in future $N_{\text{f}}$ stops. In Eq.~\eqref{prediction_eq} and Eq.~\eqref{multi_step}, the multi-step ATP is converted into the prediction of delays by the well-trained deep neural network $ f_{\pmb{\theta}}$.

Similar to~\cite{liarrivalnet}, the temporal feature $\mathbf{F}^{\text{temporal}}_{i-N_{\text{p}}+1:i}$ is the sequential combination of information in past $N_{\text{p}}$ stops,
\begin{equation}
\mathbf{F}^{\text{temporal}}_{i-N_{\text{p}}+1:i} = \{\mathbf{F}^{\text{}}_{j}\} \ \ \ \ 
 \forall j\in[i-N_{\text{p}}+1,i], \ \ j\in\mathbb{Z}
\end{equation}
\begin{equation}
\mathbf{F}^{\text{}}_{j}=[S^{\text{t}}_{j},T^{\text{t}}_{j}, T^{\text{d}}_{j}, I_{j}, \overline{T}^{\text{t}}_{j}]
\end{equation}
where $S^{\text{t}}_{j}$ and $T^{\text{t}}_{j}$ are travel distance and time from stop $j-1$ to stop $j$. $T^{\text{d}}_{j}$ is the delay at stop $j$. It is calculated from the difference of timetable and real-time arrival information. $\overline{T}^{\text{t}}_{j}$ is the average of $T^{\text{t}}_{j}$ in collected data. $I_{j}$ is the boolean value, which indicates whether a traffic signal will influence the vehicle from  stop $j-1$ to stop $j$. Whether a link connecting two stops passes through a traffic signal infrastructure is the dynamic feature, which may change temporally in a sequence. The matching process between infrastructure and daily updated public transport routes in Dresden is illustrated in \cite{liarrivalnet}. $\mathbf{F}^{\text{context}}$ consists of two binary features: peak/off peak hour and weekday/weekend. The rush hours are from from 7:00 AM to 9:00 AM and 4:00 PM to 7:00 PM on weekdays. 

\subsection{Series stationarization and over-stationarization issue}
For naturalistic data collected in the real world, non-stationarity is an intrinsic characteristic, meaning that the joint distribution of various variables changes along the time series. To improve predictability, the series stationarization is usually applied to eliminate the impact of non-stationarity. Specifically, for the temporal feature $\mathbf{F}^{\text{temporal}}_{i-N_{\text{p}}+1:i}=[\mathbf{F}^{\text{}}_{i-N_{\text{p}}+1},...,\mathbf{F}^{\text{}}_i]^\top\in \mathbb{R}^{N_{\text{p}}\times C}$ with feature length $C$, it normalizes and de-normalizes each variable along the temporal dimension. 
\begin{equation}\label{normal-3}
{\mathbf{F}^{\text{}}_{j}}^{\prime} = \frac{1}{\pmb{\sigma}_{\text{}}} \odot (\mathbf{F}^{\text{}}_j - \pmb{\mu}) \ \ \ \ 
 \forall j\in[i-N_{\text{p}}+1,i], \ \ j\in\mathbb{Z}
\end{equation}
\begin{equation}\label{normal-4}
\hat{\mathbf{T}}^{\text{d}}_{k} = \pmb{\sigma}\odot\hat{\mathbf{T}}^{\text{d}\prime}_{k}+ \pmb{\mu} \ \ \ \ 
 \forall k\in[i+1,{i+N_{\text{f}}}], \ \ k\in\mathbb{Z}
\end{equation}
where $\pmb{\mu}\in \mathbb{R}^{C\times 1}$ and $\pmb{\sigma}\in \mathbb{R}^{C\times 1}$ are mean and standard deviation of all variables in temporal dimension. $\hat{\mathbf{T}}^{\text{d}\prime}_{k}$ is the output of predictor, which is de-normalized to $\hat{\mathbf{T}}^{\text{d}}_{k}$ in the original feature space.
\begin{equation}\label{normal-1}
    \pmb{\mu} = \frac{1}{N_{\text{p}}} \sum_{j=i-N_{\text{p}}+1}^{i} \mathbf{F}^{\text{}}_j
\end{equation}
\begin{equation}\label{normal-2}
\pmb{\sigma}_{\text{}}^2 = \frac{1}{N_{\text{p}}}\sum_{j=i-N_{\text{p}}+1}^{i} (\mathbf{F}^{\text{}}_j - \pmb{\mu})^2
\end{equation}

Based on Eqs.~\eqref{normal-3}-\eqref{normal-2}, $\mathbf{F}^{\text{}}_j\in \mathbb{R}^{1\times C}$ is normalized as ${\mathbf{F}^{\text{}}_{j}}^{\prime}\in \mathbb{R}^{1\times C}$. Then, the normalized input feature ${\mathbf{F}^{\text{temporal}}_{i-N_{\text{p}}+1:i}}^{\prime}=[{\mathbf{F}^{\text{}}_{i-N_{\text{p}}+1}}^{\prime},...,{\mathbf{F}^{\text{}}_i}^{\prime}]^\top\in \mathbb{R}^{N_{\text{p}}\times C}$ is combined with $\mathbf{F}^{\text{context}}$ as  $\mathbf{F}_{\text{1D},i}\in \mathbb{R}^{N_{\text{p}}\times(C+N_{\text{c}})}$:
\begin{equation}
\mathbf{F}_{\text{1D},i}=\text{Concat}({\mathbf{F}^{\text{temporal}}_{i-N_{\text{p}}+1:i}}^{\prime},\mathbf{F}^{\text{context}})
\end{equation}
where $N_{\text{c}}$ is the feature dimension of contextual information. In order to simplify the expression, $\mathbf{F}_{\text{1D},i}$ is replaced by $\mathbf{F}_{\text{1D}}$ by removing subscript $i$. Similarly, ${\mathbf{F}^{\text{temporal}}_{i-N_{\text{p}}+1:i}}\in \mathbb{R}^{N_{\text{p}}\times C}$, ${{\mathbf{F}^{\text{temporal}}_{i-N_{\text{p}}+1:i}}}^{\prime}\in \mathbb{R}^{N_{\text{p}}\times C}$, $\hat{\mathbf{T}}^{\text{d}}_{i+1:i+N_{\text{f}}}$, $\mathbf{T}^{\text{s}}_{i+1:i+N_{\text{f}}}$ and $\hat{\mathbf{T}}^{\text{a}}_{i+1:i+N_{\text{f}}}$ are replaced by $\mathbf{F}^{\text{temporal}}$, ${\mathbf{F}^{\text{temporal}}}^{\prime}$, $\hat{\mathbf{T}}^{\text{d}}$, $\hat{\mathbf{T}}^{\text{s}}$ and $\hat{\mathbf{T}}^{\text{a}}$  by removing subscripts, respectively. Then, $\mathbf{F}_{\text{1D}}$ is sent to the multi-step predictor with output $\hat{\mathbf{T}}^{\text{d}\prime}_{k}$.

Recent studies have experimentally demonstrated that the series stationarization can enhance predictability in long-term forecasting~\cite{hyndman2018forecasting}. However, the recovery of non-stationary characteristics in the original time series  relies on the de-normalization process, which only uses the statistical features (e.g., mean and variance) in the normalization. The  procedure does not involve any learnable parameters, which cannot reflect the non-stationarity in the original time series. This leads to potential degradations  in performance, as illustrated in the following example. 

\begin{example}\label{example_1}
\textit{For two different original sequences,} $\mathbf{F}^{1,\text{temporal}}_{i-N_{\text{p}}+1:i}$ \textit{and} $\mathbf{F}^{2,\text{temporal}}_{i-N_{\text{p}}+1:i}$, \textit{the series stationarization can be simply described as:}
\begin{equation}
    {\mathbf{F}^{p,\text{temporal}}_{i-N_{\text{p}}+1:i}}^{\prime}=\pmb{\alpha}\mathbf{F}^{p,\text{temporal}}_{i-N_{\text{p}}+1:i}+\pmb{\beta}, \ \ p\in\{1,2\}
\end{equation}
\textit{where} $\pmb{\alpha}$ and $\pmb{\beta}$ \textit{are scaling and shifting parameters in normalization. After the normalization,} $\mathbf{F}^{1,\text{temporal}}_{i-N_{\text{p}}+1:i}$ \textit{and} $\mathbf{F}^{2,\text{temporal}}_{i-N_{\text{p}}+1:i}$ may become identical or similar. Physically, it generates the same output for two different inputs. In other words, the deep neural network model cannot capture the non-stationary characteristics from the normalized sequences.
\end{example}

To address this issue, this work aims to recover the influence of non-stationary effects from the original sequences within the basic predictor, as detailed in Section~\ref{ns_cnn} and Section~\ref{ns_attention}.


\begin{figure}[h!]
  \centering
  \includegraphics[width=0.6\textwidth]{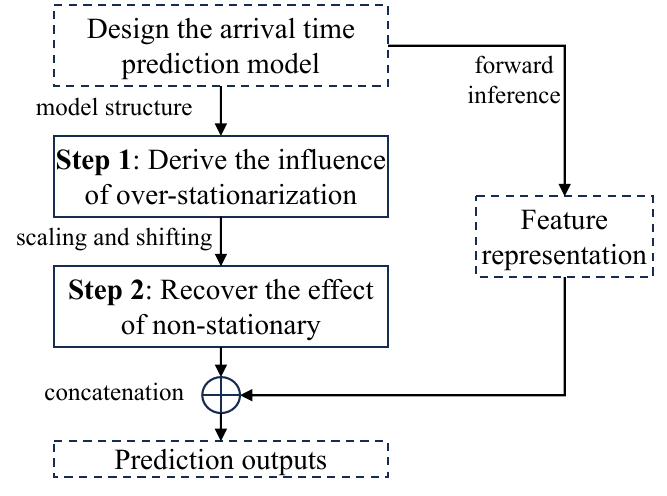}
  \caption{The illustration of non-stationary effect recovery.}
  \label{fig_NSATP_Procedure}
\end{figure}
\subsection{The procedure of non-stationary effect recovery (NSATP)}
As mentioned above, the cause of over-stationarization is the loss of non-stationarity in normalized time series, which prevents the model from capturing useful characteristics related to non-stationarity. As shown in Fig.~\ref{fig_NSATP_Procedure}, the procedure of  non-stationary effect recovery consists of two steps: the derivation of over-stationarization impact and the recovery of non-stationary effect.

\begin{itemize}
\item \textbf{Step 1}: To capture the over-stationarization, it is necessary to derive the feature representation after learning the original non-stationary time series. Therefore, the problem is converted into analyzing the impact of the original non-stationary time series on the information extraction module.
\item \textbf{Step 2}: To recover the non-stationary effect in the original time series, MLP-based neural networks are designed for element-wise scaling and shifting.
\end{itemize}

\begin{figure*}[t!]
  \centering
  \includegraphics[width=0.98\textwidth]{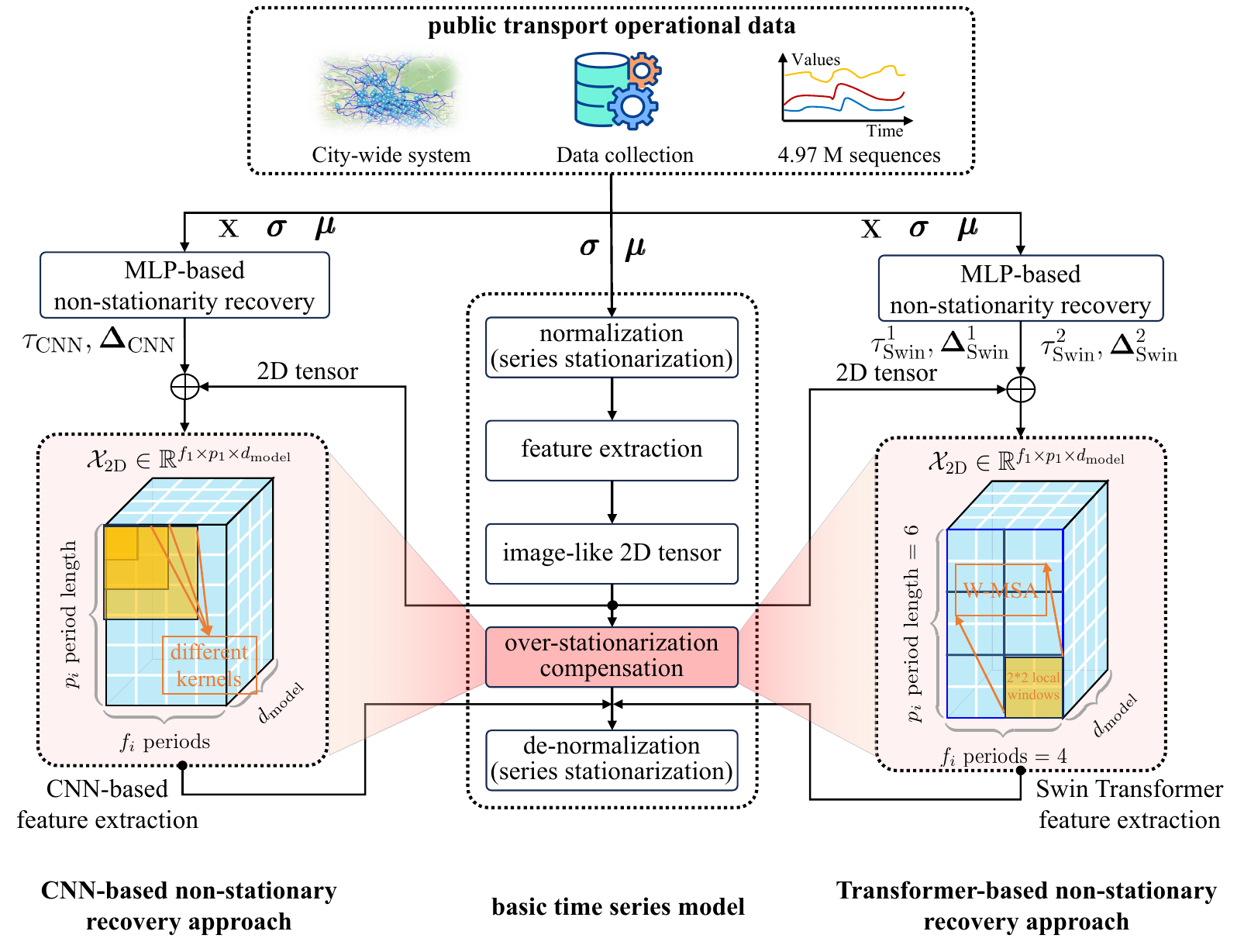}
  \caption{The overall illustration of the proposed non-stationary effect recovery method (NSATP). Top: the large-scale data collection from the public transport operational system; bottom middle: the integration of basic model and non-stationary recovery; bottom left: CNN-based non-stationary recovery; bottom right: Swin Transformer-based non-stationary recovery.}
  \label{fig_framework}
\end{figure*}

The overall illustration of the proposed non-stationary effect recovery method (NSATP) is shown in Fig.~\ref{fig_framework}. Based on the public transport operational data and the basic model, the influence of over-stationarization is analyzed and two compensation approaches are designed for corresponding variants. In Section~\ref{ns_cnn} and Section~\ref{ns_attention}, the non-stationary effects are derived and the neural network-based compensation are proposed  for CNN-based and Swin Transformer-based variants, respectively.

\section{Non-stationary effect in CNN}\label{ns_cnn}
The key in analyzing the non-stationarity effects in CNN-based \textit{ArrivalNet} is exploring whether each submodule satisfies the linearity. If satisfied, the non-stationarity of the normalized time series can be restored from the output with de-normalization; if not satisfied, it is necessary to derive the impact of the original time series in the nonlinear process.

In the CNN-based \textit{ArrivalNet}, information extraction is primarily composed of four parts: the fast Fourier transform (FFT), 1D convolutional operation, 2D convolutional operation, and nonlinear activation function between different layers.  To make consistency in expression, the input to all submodules in the following discussion is uniformly denoted as $\mathbf{X}$. To facilitate the discussion on linearity, the assumption about the standard deviation of series stationarization is made.

\begin{assumption}\label{assum_1}
\textit{Each variable in the time series shares the same standard deviation.}
\end{assumption}

Based on Assumption~\ref{assum_1}, the vector $\pmb{\sigma}\in \mathbb{R}^{C\times 1}$ is reduced as scalar. Therefore, if $f$ is a linear operation, Proposition~\ref{proposi_1} can be deduced based on Property~\ref{property_1}. The proof of Proposition~\ref{proposi_1} is detailed in Appendix~\ref{appendix_proposi_1}. 
\begin{property}\label{property_1}
\textit{A function $f$ is said to be linear if it satisfies the following two properties for all vectors $\mathbf{X}, \mathbf{Y}$ and any scalar $a \in \mathbb{R}$:
\begin{itemize}
\item Additivity: $f(\mathbf{X} + \mathbf{Y}) = f(\mathbf{X}) + f(\mathbf{Y})$.
\item Homogeneity: $f(a\mathbf{X}) = af(\mathbf{X})$.
\end{itemize}}
\end{property}

\begin{proposition}\label{proposi_1}
\textit{With the original time series $\mathbf{X} = [\mathbf{X}_1,\mathbf{X}_2,...,\mathbf{X}_{T}]^{\top}$ and normalized time series $\mathbf{X}^{\prime} = [\mathbf{X}^{\prime}_1,\mathbf{X}^{\prime}_2,...,\mathbf{X}^{\prime}_{T}]^{\top}$, their relationship is as follow:}
\begin{equation}
f(\mathbf{X})= \sigma f({\mathbf{X}}^{\prime})+\mathbf{1}\pmb{\mu}^{\top}_{f(\mathbf{X})}
\end{equation}
where $\mathbf{1}\in\mathbb{R}^{T\times 1}$ and $\sigma$ is the scalar standard deviation. $\pmb{\mu}_{f(\mathbf{X})} =  \frac{1}{T}\sum_{i=1}^{T}f(\mathbf{X}_i)$.  $T$ is the length of sequence.
\end{proposition}

The proof of linearity in the fast Fourier transform is detailed in Appendix~\ref{appendix_FFT}.

\subsection{The linearity of one dimension convolutional operation (conv1d)}
Given the input signal $\mathbf{X}$ with length $T$, the one dimension convolutional operation of the $t^{th}$ element is formulated as:
\begin{equation}
 \mathbf{Y}[t]=(\mathbf{X}*\mathbf{h})[t]= \sum_{k=0}^{M-1} \mathbf{X}[t + k] \cdot \mathbf{h}[k]
\end{equation}
where $\mathbf{h}$ is the kernel with size $M$ and $\mathbf{X}*\mathbf{h}$ donates the convolution of $\mathbf{X}$. $t+k$ accounts for the current position of the kernel over the input sequence $\mathbf{X}$. It means that for each output value $\mathbf{Y}[t]$ only $M$ elements currently being covered by the kernel $\mathbf{h}$ is considered. In the convolutional operation, the kernel $\mathbf{h}$ slides over the one dimensional sequence $\mathbf{X}$ and  calculate the sum of the element-wise products. The kernel is shifted by one element each time, from the start to the end of the input sequence $\mathbf{X}$. To prove the additivity, given two input signals $\mathbf{X}_1$ and $\mathbf{X}_2$, the $t^{th}$ element of $\mathbf{Y}_1 + \mathbf{Y}_2$ is formulated as:
\begin{equation}
\begin{aligned}
&(\mathbf{Y}_1 + \mathbf{Y}_2)[t]\\
&=\sum_{k=0}^{M-1} (\mathbf{X}_1[t + k] + \mathbf{X}_2[t + k]) \cdot \mathbf{h}[k] \\
&= \sum_{k=0}^{M-1} \mathbf{X}_1[t + k] \cdot \mathbf{h}[k] + \sum_{k=0}^{M-1} \mathbf{X}_2[t + k] \cdot \mathbf{h}[k]\\
&= (\mathbf{X}_1*\mathbf{h})[t] + (\mathbf{X}_2*\mathbf{h})[t] \  \  \  \  t\in [1, T], t\in\mathbb{Z}
\end{aligned}
\end{equation}

To prove the homogeneity, give the input signal $\mathbf{X}$ and scalar $a$, we have 
\begin{equation}
\begin{aligned}
(a\mathbf{X}*\mathbf{h})[t] &=\sum_{k=0}^{M-1} (a \mathbf{X}[t + k]) \cdot \mathbf{h}[k] \\
&= a \sum_{k=0}^{M-1} \mathbf{X}[t + k] \cdot \mathbf{h}[k] \\
&= a(\mathbf{X}*\mathbf{h})[t]  \  \  \  \  t\in [1, T], t\in\mathbb{Z}
\end{aligned}
\end{equation}

Finally, the additivity and homogeneity are proved. In \textit{ArrivalNet}, the one dimensional convolution is applied in the feature embedding process. Based on Proposition~\ref{proposi_1}, the embedding of normalized $\mathbf{X}^{\prime}$ is expressed as follows:
\begin{equation}
    f({\mathbf{X}}^{\prime})= \frac{1}{\sigma}(f_{\text{embed}}(\mathbf{X})-\mathbf{1}\pmb{\mu}^{\top}_{f_{\text{embed}}(\mathbf{X})})
\end{equation}
where $\pmb{\mu}_{f_{\text{embed}}(\mathbf{X})}\in \mathbb{R}^{d_{\text{model}}\times 1}$ is the mean of one dimensional convolution-based embedding function and $\mathbf{X}$ is the original sequence.

\subsection{The linearity of two dimension convolutional operation (conv2d)}
In CNN, a two-dimensional convolution (2D convolution) involves sliding a two-dimensional kernel  over a two-dimensional input (e.g., image) and computing the output as the sum of element-wise products. With the input signal $\mathbf{X}\in\mathbb{R}^{H\times W}$, the 2D convolution of $(i,j)^{th}$ element is defined as:
\begin{equation}
    \mathbf{Y}[i, j] = (\mathbf{X} * \mathbf{h})[i, j] = \sum_{m=0}^{M-1} \sum_{n=0}^{N-1} \mathbf{X}[i + m, j + n] \cdot \mathbf{h}[m, n]
\end{equation}
where $M$ and $N$ are the height and width of the 2D convolution kernel. The derivation of linearity in two dimension convolutional operation is an extension of 1D convolutional operation.  Given two input signals $\mathbf{X}_1$ and $\mathbf{X}_2$, the $(i,j)^{th}$ element of $\mathbf{Y}_1 + \mathbf{Y}_2$ is formulated as:
\begin{equation}
    \begin{aligned}
        & \mathbf{Y}_1[i,j] + \mathbf{Y}_2[i,j] = (\mathbf{X}_1 + \mathbf{X}_2) * \mathbf{h}[i, j] \\
        &=  \sum_{m=0}^{M-1} \sum_{n=0}^{N-1}  (\mathbf{X}_1[i + m, j + n] + \mathbf{X}_2[i + m, j + n]) \cdot \mathbf{h}[m, n]\\
        &=  \sum_{m=0}^{M-1} \sum_{n=0}^{N-1}  \mathbf{X}_1[i + m, j + n] \cdot \mathbf{h}[m, n] \\
        &\ \ \ \ +  \sum_{m=0}^{M-1} \sum_{n=0}^{N-1}  \mathbf{X}_2[i + m, j + n] \cdot \mathbf{h}[m, n]\\
        &= (\mathbf{X}_1 * \mathbf{h})[i, j] + (\mathbf{X}_2 * \mathbf{h})[i, j]
    \end{aligned}
\end{equation}

Similarly, the homogeneity is derived as follows:
\begin{equation}
    \begin{aligned}
        &(a \cdot \mathbf{X}) * \mathbf{h}[i, j] \\
        &= \sum_{m=0}^{M-1} \sum_{n=0}^{N-1} (a \cdot \mathbf{X}[i + m, j + n]) \cdot \mathbf{h}[m, n]\\
        &= a \cdot\sum_{m=0}^{M-1} \sum_{n=0}^{N-1} \mathbf{X}[i + m, j + n] \cdot \mathbf{h}[m, n]\\
        &= a \cdot (\mathbf{X} * \mathbf{h})[i, j]
    \end{aligned}
\end{equation}

Finally, the additivity and homogeneity are proved in 2D convolutional operation. In \textit{ArrivalNet}, the 2D convolution is applied on the 2D image-like tensor. Based on Proposition~\ref{proposi_1}, each layer of 2D convolution-based neural network $f_{\text{Conv2D}}$ has the following formulation:
\begin{equation}
    f({\mathbf{X}}^{\prime})= \frac{1}{\sigma}(f_{\text{Conv2D}}(\mathbf{X})-\mathbf{1}\pmb{\mu}^{\top}_{f_{\text{Conv2D}}(\mathbf{X})})
\end{equation}
where $\mathbf{X}$ and $\mathbf{X}^{\prime}$ are the original and normalized sequences.  $\pmb{\mu}_{f_{\text{Conv2D}}(\mathbf{X})}\in \mathbb{R}^{d_{\text{model}}\times 1}$ is the mean of 2D convolutional function outputs. 

\subsection{The influence of activation function}
For the DNN architecture composed of multiple CNNs, it is necessary to use nonlinear activation functions between different CNN layers to enhance the model's nonlinearity. In the basic predictor, the rectified linear unit (ReLU) function is employed, which is also the most commonly used activation function in DNN. With the image-like two dimensional tensor input $\mathbf{X}\in\mathbb{R}^{f\times p\times d_{\text{model}}}$, ReLU can be expressed as:
\begin{equation}
    \text{ReLU}(\mathbf{X})=\max(0,\mathbf{X})
\end{equation}
where $f$ and $p$ are frequency and periodic length from the FFT of the time series, respectively. $d_{\text{model}}$ is the model space. Based on Assumption~\ref{assum_1}, all elements in $\mathbf{X}$ share the same standard deviation. With ${\mathbf{X}}^{\prime}= \frac{1}{\sigma}((\mathbf{X})-\pmb{\mu}_{f_{\text{Conv2D}}(\mathbf{X})})$, 
\begin{equation}
    \text{ReLU}(\mathbf{X}))=\max(0, \sigma\mathbf{X}^{\prime}+\pmb{\mu}_{f_{\text{Conv2D}}(\mathbf{X})})
\end{equation}
where $\sigma\in\mathbb{R}^{+}$ and $\pmb{\mu}_{f_{\text{Conv2D}}(\mathbf{X})}\in \mathbb{R}^{d_{\text{model}}\times 1}$ are the standard deviation for all elements and mean for each model space. Then, the relationship between $\mathbf{X}$ and $\mathbf{X}^{\prime}$ is as follow:
\begin{equation}
    \text{ReLU}(\mathbf{X}) = \sigma\max\left(-\frac{ \pmb{\mu}_{f_{\text{Conv2D}}(\mathbf{X})}}{\sigma}, \mathbf{X}^{\prime}\right) +  \pmb{\mu}_{f_{\text{Conv2D}}(\mathbf{X})}
\end{equation}

\begin{figure}[t]
  \centering
  \includegraphics[width=0.5\textwidth]{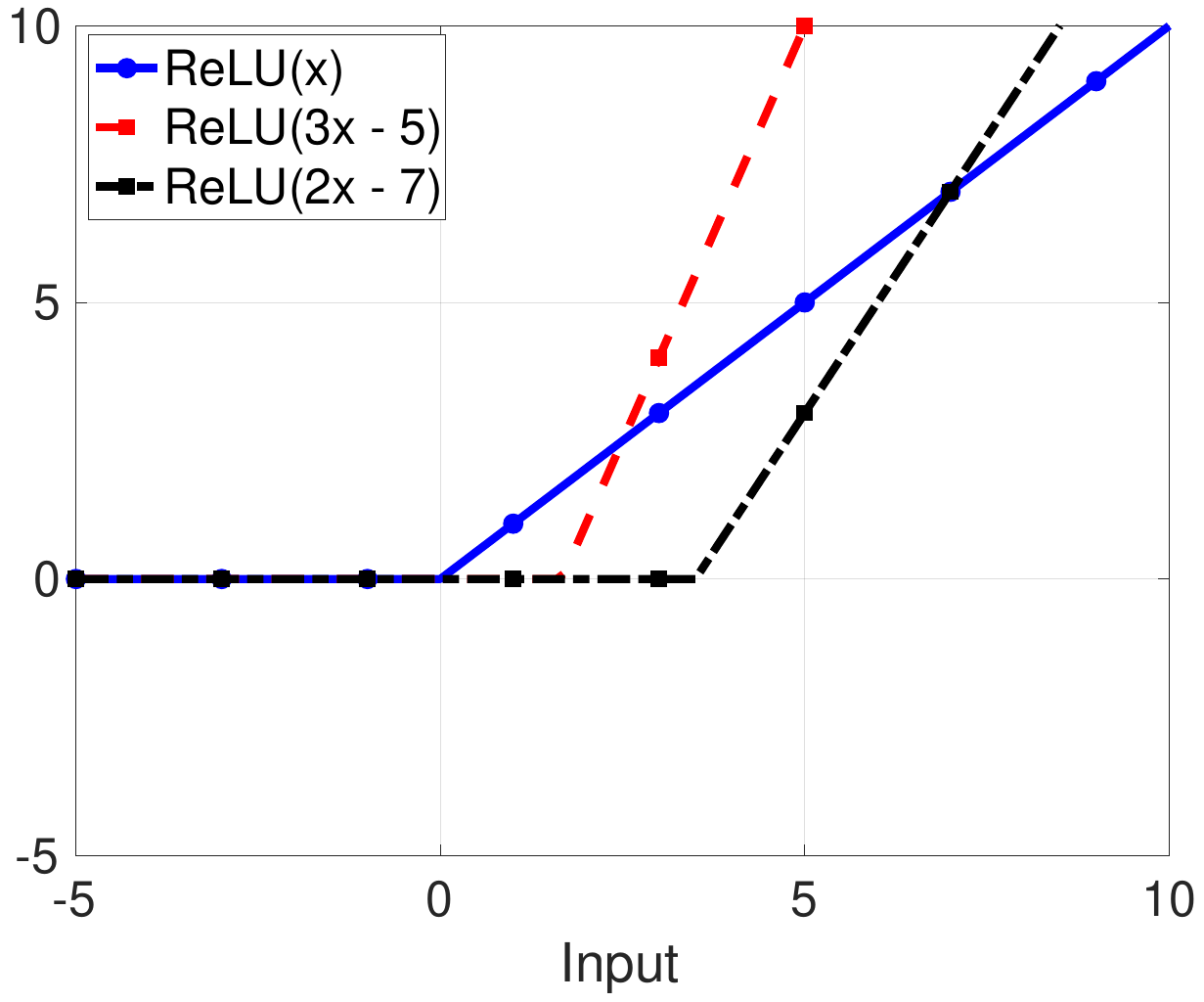}
  \caption{The visualization of ReLU and its variants.}
  \label{fig:relu}
\end{figure}
As illustrated in Fig.~\ref{fig:relu}, comparing $\text{ReLU}(x)$ with $\text{ReLU}(3x-5)$ and $\text{ReLU}(2x-7)$, it involves scaling and shifting the ReLU function with factor $\sigma$ and $\pmb{\mu}_{f_{\text{Conv2D}}(\mathbf{X})}$, respectively. Additionally, set values less than \(-\pmb{\mu}_{f_{\text{Conv2D}}(\mathbf{X})}/\sigma\) to zero as the final step.

\subsection{Non-stationary effect recovery for CNN}\label{compensation_cnn}
\subsubsection{The principle non-stationary effect recovery}
From the analysis above, to recover the non-stationarity in the original sequence, the influence of standard deviation $\sigma$, $\mu_{f_{\text{embed}}(\mathbf{X})}$ in the one dimensional CNN, $\mu_{f_{\text{Conv2D}}(\mathbf{X})}$ in the two dimensional CNN, and the nonlinear activation function need to be estimated. The compensation of the over-stationarization is formulated as the scaling and shifting of the elements within the image-like 2D tensor $\mathbf{X}\in\mathbb{R}^{f\times p\times d_{\text{model}}}$, which can be transformed into $\mathbf{X}\in\mathbb{R}^{T\times d_{\text{model}}}$ with $T=f\times p$.
\subsubsection{Compensation design}
Since the time series sent to CNN is the normalized sequence, the trained weights in $f_{\text{embed}}$ and $f_{\text{Conv2D}}$ are for the normalized input feature. It is impossible to directly calculate $\pmb{\mu}_{f_{\text{FFT}}(\mathbf{X})}$,  $\pmb{\mu}_{f_{\text{embed}}(\mathbf{X})}$ and $\pmb{\mu}_{f_{\text{Conv2D}}(\mathbf{X})}$ in the original sequence. To recover the implicit non-stationarity information, we try to learn the compensation of over-stationarization from the original sequence. Additionally, reducing vectorized standard deviation to a scalar involves estimating the scalarized $\sigma$, which can be learned from the original sequence and shared for all elements. Similarly, the compensation for $\pmb{\mu}_{f_{\text{embed}}(\mathbf{X})}$ and $\pmb{\mu}_{f_{\text{Conv2D}}(\mathbf{X})}$ in  one-dimensional and two-dimensional CNNs are vectorized tensors, which repeated $T$ times for the whole sequence. 

\begin{figure*}[t!]
  \centering
  \includegraphics[width=0.8\textwidth]{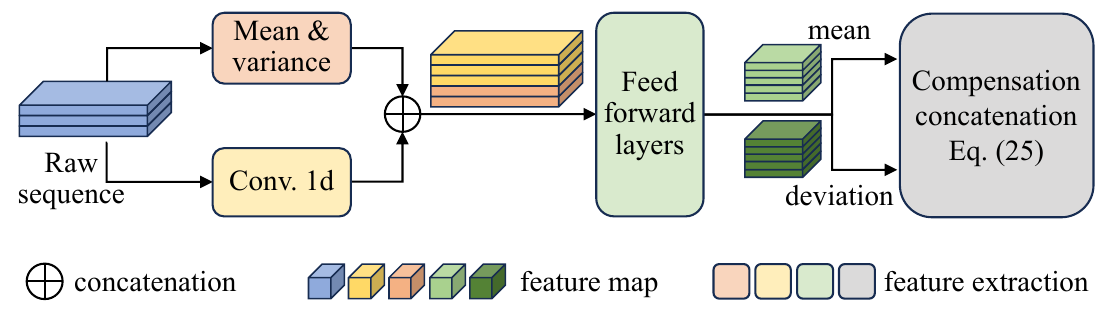}
  \caption{The MLP-based neural network in Eqs.~\eqref{cnn_factor_1} and~\eqref{cnn_factor_2}.}
  \label{fig_MLP}
\end{figure*}

Besides considering the influence of scaling and shifting in convolutional operations, $\text{ReLU}(\mathbf{X})$ is also affected by the corresponding element values. It means the shifting compensation should be element-wise. Overall, to eliminate the over-stationarization in the CNN-based \textit{ArrivalNet}, a scalar scaling factor $\tau_{\text{CNN}}\in\mathbb{R^{+}}$ and an element-wise shifting factor $\mathbf{\Delta}_{\text{CNN}}\in\mathbb{R}^{T\times d_{\text{model}}}$ are required. Inspired by~\cite{liu2022non}, two \text{MLP}-based neural networks are designed for non-stationary recovery.
\begin{equation}\label{cnn_factor_1}
    \log \tau_{\text{CNN}} = \text{MLP}(\pmb{\sigma}_{\mathbf{X}},\mathbf{X})
\end{equation}
\begin{equation}\label{cnn_factor_2}
    \mathbf{\Delta}_{\text{CNN}} = \text{MLP}(\pmb{\mu}_{\mathbf{X}},\mathbf{X})
\end{equation}
where $\pmb{\sigma}_{\mathbf{X}}\in \mathbb{R}^{C\times 1}$ and $\pmb{\mu}_{\mathbf{X}}\in \mathbb{R}^{C\times 1}$ are mean and standard deviation of the original sequence. The architecture of \text{MLP}-based neural network is shown in Fig.~\ref{fig_MLP}. 

\subsubsection{Compensation concatenation}
\begin{figure}[t!]
  \centering
  \includegraphics[width=0.55\textwidth]{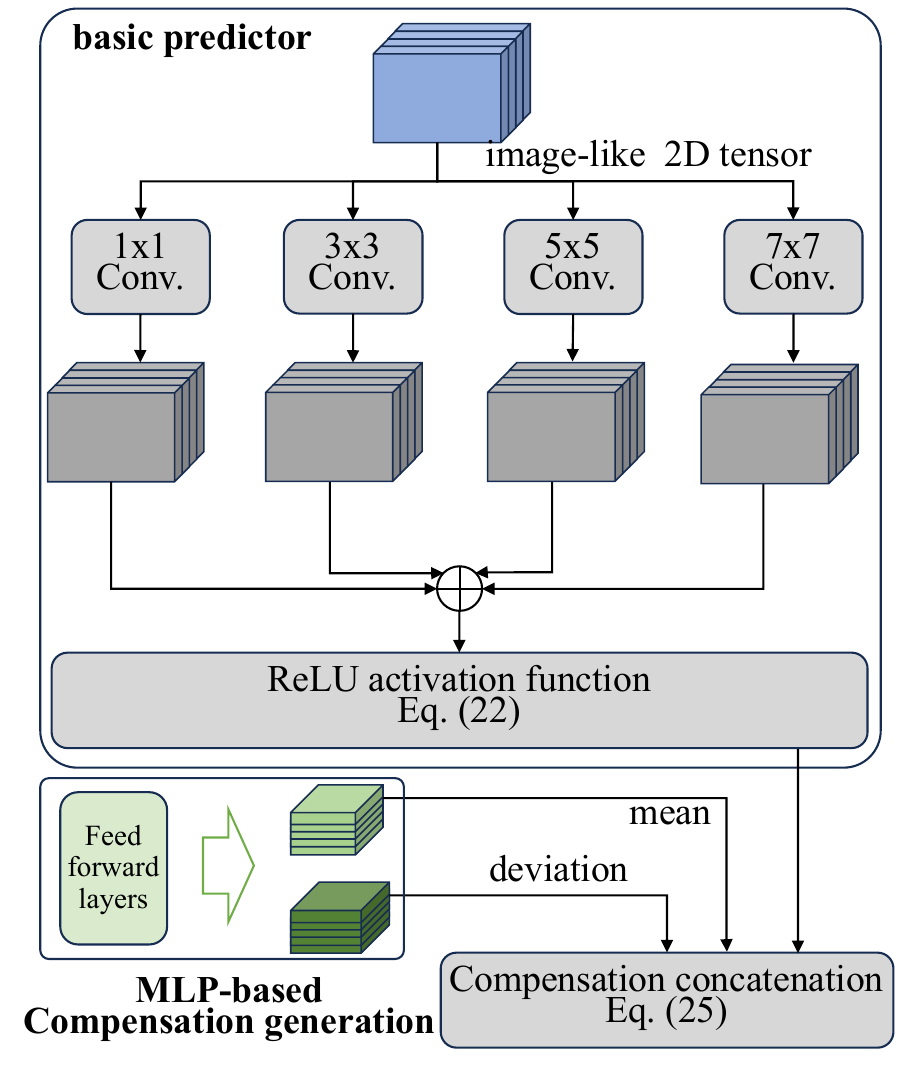}
  \caption{The CNN-based feature extraction for 2D image-like tensor with non-stationary effect recovery in Eq.~\eqref{cnn_compensation}: NSATP-1 (CNN). The meaning of elements is consistent with Fig.~\ref{fig_MLP}.}
  \label{fig_cnn_illustration}
\end{figure}

Based on Eq.~\eqref{cnn_factor_1} and Eq.~\eqref{cnn_factor_2}, the compensation is applied at the end of each 2D block,
\begin{equation}\label{cnn_compensation}
    \mathbf{X}^{c}_{\text{CNN}} = \tau_{\text{CNN}}\mathbf{X}^{\prime}+  \mathbf{\Delta}_{\text{CNN}}
\end{equation}
where $\mathbf{X}^{c}_{\text{CNN}}$ is the tensor with over-stationarization compensation. Fig.~\ref{fig_cnn_illustration} shows the integration of  compensation and CNN-based \textit{ArrivalNet}. The information extraction flow of the proposed NSATP-1 (CNN) is detailed in Algorithm~\ref{algorithm_cnn}.

\begin{algorithm*}
\caption{NSATP-1 (CNN)}
\label{algorithm_cnn}
\KwIn{historical input length $N_{\text{p}}$, predicted output length $N_{\text{f}}$, model space $d_{\text{model}}$, number of 2D block $L$, selected frequencies in FFT $k$, temporal feature $\mathbf{F}^{\text{temporal}}\in \mathbb{R}^{N_{\text{p}}\times C}$, contextual feature $\mathbf{F}^{\text{context}}\in \mathbb{R}^{N_{\text{p}}\times N_{\text{c}}}$, scheduled arrival time ${\mathbf{T}}^{\text{s}}\in\mathbb{R}^{N_{\text{f}}\times 1}$.}
\KwOut{arrival time $\hat{\mathbf{T}}^{\text{a}}$.}
$\mathbf{X}^{\prime}, \pmb{\mu}_{\mathbf{X}}, \pmb{\sigma}_{\mathbf{X}} = \text{Normalization}(\mathbf{F}^{\text{temporal}})$ as in Eqs.~\eqref{normal-3}-\eqref{normal-2}. 
\hfill $\triangleright$ $\mathbf{X}^{\prime}\in \mathbb{R}^{N_{\text{p}}\times C}$,  $\pmb{\mu}\in \mathbb{R}^{C\times 1}$, $\pmb{\sigma}\in \mathbb{R}^{C\times 1}$\\
Estimate $\tau_{\text{CNN}}$, $\mathbf{\Delta}_{\text{CNN}}$ by MLP as in Eqs.~\eqref{cnn_factor_1}-\eqref{cnn_factor_2}.
\hfill $\triangleright$ $\tau_{\text{CNN}}\in\mathbb{R^{+}}$, $\mathbf{\Delta}_{\text{CNN}}\in\mathbb{R}^{T\times d_{\text{model}}}$\\

$\hat{\mathbf{X}}_{\text{1D}}^{0} = \text{Embed}\left(\text{Concatenate}\left(\mathbf{X}^{\prime},\mathbf{F}^{\text{context}}\right)\right)$ by Conv1d-based feature embedding. 
\hfill $\triangleright$ $\mathbf{X}_{\text{1D}}^{0}\in\mathbb{R}^{T\times d_{\text{model}}}$, $T=N_{\text{p}}+N_{\text{f}}$\\

\For{$l \in \left\{1,2,..., L \right\}$}{
1D$\rightarrow$2D $(\hat{\mathbf{X}}_{\text{2D},1}^{l-1},...,\hat{\mathbf{X}}_{\text{2D},k}^{l-1}),(\mathbf{A}_{f_{1}}^{l-1},...,\mathbf{A}_{f_{k}}^{l-1})=\text{FFT}(\hat{\mathbf{X}}_{\text{1D}}^{l-1})$. 
\hfill $\triangleright$$\mathbf{X}_{\text{2D},i}^{l-1} \in \mathbb{R}^{f_{i}\times p_{i}\times d_{\text{model}}}\ \ \forall i\in [1,k], i\in \mathbb{Z}$ \\

\For{$i \in \left\{1,2,..., k\right\}$}{
$\hat{\mathbf{X}}_{\text{2D},i}^{l}=\text{Inception}_{\text{conv2d}}(\hat{\mathbf{X}}_{\text{2D},i}^{l-1})$.
\hfill $\triangleright$ CNN-based vision backbone, $\mathbf{X}_{\text{2D},i}^{l-1} \in \mathbb{R}^{f_{i}\times p_{i}\times d_{\text{model}}}$\\
$\hat{\mathbf{X}}_{\text{1D},i}^{l}=\text{Reshape}(\hat{\mathbf{X}}_{\text{2D},i}^{l})+\hat{\mathbf{X}}_{\text{1D},i}^{l-1}$.
\hfill $\triangleright$ residual connection, $\mathbf{X}_{\text{1D},i}^{l}\in\mathbb{R}^{T\times d_{\text{model}}}$
}
$\hat{\mathbf{A}}_{\text{Top k}} =\text{Softmax}(\mathbf{A}_{\text{Top k}})$. \hfill $\triangleright$ rescale to $[0,1]$\\
$\hat{\mathbf{X}}_{\text{1D}}^{l} =\sum_{i=1}^{k} \hat{\mathbf{A}}_{f_{i}} \times \hat{\mathbf{X}}_{\text{1D}}^{l}$.
\hfill $\triangleright$  weighted sum, $\mathbf{X}_{\text{1D}}^{l}\in\mathbb{R}^{T\times d_{\text{model}}}$\\
}
Compensation $\mathbf{X}^{c}_{\text{CNN}} = \tau_{\text{CNN}}\hat{\mathbf{X}}_{\text{1D}}^{l}+\mathbf{\Delta}_{\text{CNN}}$ as in Eq.~\eqref{cnn_compensation}. \hfill $\triangleright$  Non-stationary effect recovery\\

Calculate delays $\hat{\mathbf{T}}^{\text{d}}=\text{Trun}(\underset{d_{\text{model}}\rightarrow 1}{\text{Linear}}(\mathbf{X}^{c}_{\text{CNN}}))$.
\hfill $\triangleright$ truncation and linear projection back, $\hat{\mathbf{T}}^{\text{d}}\in\mathbb{R}^{N_{\text{f}}\times 1}$\\

Arrival time $\hat{\mathbf{T}}^{\text{a}}=\text{de-normalization}(\hat{\mathbf{T}}^{\text{d}})+{\mathbf{T}}^{\text{s}}$ as in Eq.~\eqref{multi_step}.
\hfill $\triangleright$  de-normalization and calculate ATP, $\hat{\mathbf{T}}^{\text{a}}\in\mathbb{R}^{N_{\text{f}}\times 1}$\\
\Return $\hat{\mathbf{T}}^{\text{a}}$
\end{algorithm*}

\section{Non-stationary effect in Swin Transformer}\label{ns_attention}
In the basic model, another variant involves extracting features from image-like two-dimensional tensors by Swin Transformer, which is a backbone model in computer vision~\cite{liu2021swin}. It divides an image-like tensor into multiple small windows and applies an attention mechanism within these windows. To avoid only capturing local information within individual windows, the Swin Transformer incorporates a shifting window operation, which establishes connections between adjacent pixels across different windows. 
\subsection{The plain self-attention mechanism}
The core structure of the Swin Transformer is the attention mechanism, which is illustrated as follows:
\begin{equation}\label{plain_attention}
        \text{Attn}(\mathbf{Q}, \mathbf{K}, \mathbf{V}) = \text{Softmax}\left(\frac{\mathbf{Q}\mathbf{K}^{\top}}{\sqrt{d_k}}\right)\mathbf{V}
\end{equation}
with $\mathbf{Q} = \mathbf{X}\mathbf{W}_{Q}$, $\mathbf{K} = \mathbf{X}\mathbf{W}_{K}$ and $\mathbf{V} = \mathbf{X}\mathbf{W}_{V}$. In Eq.~\eqref{plain_attention}, $\mathbf{X}=[\mathbf{X}_1,\mathbf{X}_2,...,\mathbf{X}_T]^{\top}\in\mathbb{R}^{T\times d_{\text{model}}}$ is the $T$-length sequence after feature embedding. $d_{\text{model}}$ is the feature space. $d_k$ is the scaling factor. $\mathbf{W}_{Q}\in\mathbb{R}^{d_{\text{model}}\times d_{k}}$, $\mathbf{W}_{K}\in\mathbb{R}^{d_{\text{model}}\times d_{k}}$ and $\mathbf{W}_{V}\in\mathbb{R}^{d_{\text{model}}\times d_{k}}$ are weights in the linear projection from the embedding model space t$d_{\text{model}}$ to the attention feature space $d_{k}$. Based on the learnable transformation, query $\mathbf{Q}\in\mathbb{R}^{T\times d_{k}}$, key $\mathbf{K}\in\mathbb{R}^{T\times d_{k}}$, and value $\mathbf{V}\in\mathbb{R}^{T\times d_{k}}$ are calculated. \text{Softmax} is the operation on each row. 

\subsection{The influence of series stationarization in self-attention mechanism}
Based on the Proposition~\ref{proposi_1}, the relationship of original and normalized sequences in the query, key and value functions are formulated:
\begin{equation}
\begin{aligned}
    \mathbf{Q}^{\prime} &= \left[f_{\text{query}}(\mathbf{X}^{\prime}_{1}), \ldots, f_{\text{query}}(\mathbf{X}^{\prime}_{T}),\right]^{\top} = \frac{(\mathbf{Q} - \mathbf{1} \pmb{\mu}^{\top}_{f_{\text{query}}})}{\mathbf{\sigma}}\\
    \mathbf{K}^{\prime} &= \left[f_{\text{key}}(\mathbf{X}^{\prime}_{1}), \ldots, f_{\text{key}}(\mathbf{X}^{\prime}_{T}) \right]^{\top} = \frac{(\mathbf{K} - \mathbf{1} \pmb{\mu}^{\top}_{f_{\text{key}}})}{\mathbf{\sigma}}\\
    \mathbf{V}^{\prime} &= \left[f_{\text{value}}(\mathbf{X}^{\prime}_{1}), \ldots, f_{\text{value}}(\mathbf{X}^{\prime}_{T}) \right]^{\top} = \frac{(\mathbf{V} - \mathbf{1} \pmb{\mu}^{\top}_{f_{\text{value}}})}{\mathbf{\sigma}}
\end{aligned}
\end{equation}
where $\mathbf{Q}^{\prime}$, $\mathbf{K}^{\prime}$ and $\mathbf{V}^{\prime}$ are the query, key and value for the normalized sequence, respectively. $\mathbf{1}\in\mathbb{R}^{T\times 1}$ is the all-ones vector. $\pmb{\mu}_{f_{\text{query}}}\in\mathbb{R}^{d_{k}\times 1}$, $\pmb{\mu}_{f_{\text{key}}}\in\mathbb{R}^{d_{k}\times 1}$ and $\pmb{\mu}_{f_{\text{value}}}\in\mathbb{R}^{d_{k}\times 1}$ are the mean of query, key and value functions based on the original sequence. $f_{\text{query}}$, $f_{\text{key}}$ and $f_{\text{value}}$ are the linearly projecting function equipped with learnable weights $\mathbf{W}_{Q}$, $\mathbf{W}_{K}$ and $\mathbf{W}_{V}$. To recover the hidden useful information in the original sequence $\mathbf{X}$, the relationship of the self-attention score with and without the series stationarization is derived as follows:
\begin{equation}\label{attention_soft}
\text{Softmax}\left(\frac{\mathbf{Q}\mathbf{K}^{\top}}{\sqrt{d_k}}\right)=\\
\text{Softmax}\left(\frac{\sigma^2  \mathbf{Q}^{\prime}{\mathbf{K}^{\prime}}^{\top} + \mathbf{1}(\pmb{\mu}_{f_{\text{query}}}^{\top}\mathbf{K}^{\top}) + (\mathbf{Q}\pmb{\mu}_{f_{\text{key}}})\mathbf{1}^{\top} - \mathbf{1}(\pmb{\mu}_{f_{\text{query}}}^{\top}\pmb{\mu}_{f_{\text{key}}})\mathbf{1}^{\top}}{\sqrt{d_k}}\right)  
\end{equation}
\noindent where $(\mathbf{Q}\pmb{\mu}_{f_{\text{key}}})\mathbf{1}^{\top}$ is the repeated operation of $\mathbf{Q}\pmb{\mu}_{f_{\text{key}}}\in\mathbb{R}^{T\times 1}$ by $\mathbf{1}^{\top}\in\mathbb{R}^{1\times T}$. $\pmb{\mu}_{f_{\text{query}}}^{\top}\pmb{\mu}_{f_{\text{key}}}\in\mathbb{R}$ is the scalar, which is repeated for each row and column in $\mathbf{1}(\pmb{\mu}_{f_{\text{query}}}^{\top}\pmb{\mu}_{f_{\text{key}}})\mathbf{1}^{\top}$. Since the \text{Softmax} function is the operation on each row, the attention score in Eq.~\eqref{attention_soft} is simplified as:
\begin{equation}\label{attention_soft_reduced}
\begin{aligned}
&\text{Softmax}\left(\frac{\mathbf{Q}\mathbf{K}^{\top}}{\sqrt{d_k}}\right)=\text{Softmax}\left(\frac{\sigma^2  \mathbf{Q}^{\prime}{\mathbf{K}^{\prime}}^{\top} + \mathbf{1}(\pmb{\mu}_{f_{\text{query}}}^{\top}\mathbf{K}^{\top})}{\sqrt{d_k}}\right)  
\end{aligned}
\end{equation}
In Eq.~\eqref{attention_soft_reduced}, $\mathbf{Q}\mathbf{K}^{\top}$ is influenced not only by $\mathbf{Q}^{\prime}{\mathbf{K}^{\prime}}^{\top}$ but also by $\pmb{\mu}_{f_{\text{query}}}^{\top}\mathbf{K}^{\top}$. However, the features actually received by the Swin Transformer are normalized by series stationarization and the trained weights are for $\mathbf{X}^{\prime}$, hence $\pmb{\mu}_{f_{\text{query}}}$ and $\mathbf{K}^{\top}$ cannot be directly computed. Additionally, \( \sigma \) is reduced from vector to scalar under Assumption~\ref{assum_1}, which is required to be  estimated. The compensation of over-stationarization within the \text{Softmax} is detailed in Section~\ref{compensation_swin}.

\subsection{The proof of derivation from one layer to multi-layers in Transformer}
In the above section, the influence of series stationarization within the \text{Softmax} is analysed. However, in the attention mechanism, besides calculating the attention score, the value matrices is also involved Eq.~\eqref{plain_attention}. Here we present a the proof of derivation from one layer to multi-layers in Transformer by analysing the linear property in the attention calcuation.

If we consider attention as a function of $\mathbf{V}=[\mathbf{V}_1,\mathbf{V}_2,...,\mathbf{V}_T]^{\top}\in\mathbb{R}^{T\times d_{\text{k}}}$, and represent the output of the \text{Softmax} operation as $\mathbf{E}=[\mathbf{E}_1,\mathbf{E}_2,...,\mathbf{E}_T]^{\top}\in\mathbb{R}^{T\times d_{\text{k}}}$, the attention mechanism is as follows:
\begin{equation}
 \mathbf{E} = \mathbf{\omega}\mathbf{V}
\end{equation}
where $\mathbf{\omega}=\text{Softmax}\left(\mathbf{Q}\mathbf{K}^{\top}/\sqrt{d_k}\right)\in\mathbb{R}^{T\times T}$ and the sum of each row is 1. For $\forall i\in[1,T]~i\in\mathbb{Z}$, the attention can be simplified as follows:
\begin{equation}\label{multi_layer}
\mathbf{E}_i = f_{\mathbf{\omega}}(\mathbf{V})=\sum_{j=1}^T w_{i,j} \mathbf{V}_j  
\end{equation}
where $\sum_{j=1}^T w_{i,j} = 1$ and $w_{i,j} \geq 0$. $f_{\omega}(\cdot)$ is the function for attention calculation based on $\mathbf{\omega}$. In Eq.~\eqref{multi_layer}, $f_{\omega}(\cdot)$ holds the linear property according to Property~\ref{property_1}:
\begin{equation}
f_{\omega}(a\mathbf{V}^{A}+b\mathbf{V}^{B}) = af_{\omega}(\mathbf{V}^{A})+bf_{\omega}(\mathbf{V}^{B})
\end{equation}
where $a$ and $b$ are scalar. $\mathbf{V}^{A}=[\mathbf{V}^{A}_1,\mathbf{V}^{A}_2,...,\mathbf{V}^{A}_T]^{\top}$ and $\mathbf{V}^{B}=[\mathbf{V}^{B}_1,\mathbf{V}^{B}_2,...,\mathbf{V}^{B}_T]^{\top}$ are two \textit{value} features. In the complete attention mechanism, if the non-stationary effect  within the \text{Softmax} can be well recovered, then the attention mechanism's output relative to its input becomes linear. Then, in Swin Transformer, the output of the attention mechanism is used as the input for subsequent blocks. These modules include a feed forward neural network of multiple MLPs, which naturally hold the linear property. Thus, Eq.~\eqref{attention_soft_reduced} applies across all Swin Transformer modules.

\subsection{Non-stationary effect recovery for attention}\label{compensation_swin}
\subsubsection{Compensation design}
Similar to the CNN-based \textit{ArrivalNet}, the non-stationarity recovery approach is designed for the Swin Transformer-based variant. To compensate for the influence of $\sigma^2$ and $\mathbf{K}\pmb{\mu}_{f_{\text{query}}}$ within the \text{Softmax}, a scalar scaling factor  $\tau^{1}_{\text{Swin}}=\sigma^2\in\mathbb{R^{+}}$ and a shifting factor $\mathbf{\Delta}^{1}_{\text{Swin}}=\mathbf{K}\pmb{\mu}_{f_{\text{query}}}\in\mathbb{R}^{T\times 1}$ for each model space are designed based on MLPs:
\begin{equation}\label{swin_factor_1}
    \log \tau^{1}_{\text{Swin}} = \text{MLP}(\mathbf{\sigma}_{\mathbf{X}},\mathbf{X})
\end{equation}
\begin{equation}\label{swin_factor_2}
    \mathbf{\Delta}^{1}_{\text{Swin}} = \text{MLP}(\pmb{\mu}_{\mathbf{X}},\mathbf{X})
\end{equation}
Based on Eq.~\eqref{swin_factor_1} and Eq.~\eqref{swin_factor_2}, the attention mechanism is converted to:
\begin{equation}\label{swin_compensation_1}
\begin{aligned}
\text{Attn}(\mathbf{Q}^{\prime}, &\mathbf{K}^{\prime}, \mathbf{V}^{\prime}, \tau^{1}_{\text{Swin}}, \mathbf{\Delta}^{1}_{\text{Swin}}) = \\
&\text{Softmax}\left(\frac{\tau^{1}_{\text{Swin}} \mathbf{Q}^{\prime}{\mathbf{K}^{\prime}}^{\top} + \mathbf{1} {\mathbf{\Delta}^{1}_{\text{Swin}}}^{\top}}{\sqrt{d_k}}\right) \mathbf{V}^{\prime}
\end{aligned}
\end{equation}

In Swin Transformer, $T$ depends on the window size and the patch size. In this work, the patch size is set to 1 and the length of window is 3. Therefore, $T=3*3=9$. With Eq.~\eqref{swin_compensation_1}, the non-stationarity effect within the \text{Softmax} is compensated. A complete Swin Transformer also needs to consider the impact of other modules that hold linear properties. It is similar to the compensation for 1D and 2D convolutions in the CNN-based \textit{ArrivalNet}. Therefore, after the Swin Transformer-based information extraction from image-like two-dimensional tensors, two additional scaling and shifting factors are applied:
\begin{equation}\label{swin_ff_factor_1}
    \log \tau^{2}_{\text{Swin}} = \text{MLP}(\mathbf{\sigma}_{\mathbf{X}},\mathbf{X})
\end{equation}
\begin{equation}\label{swin_ff_factor_2}
    \mathbf{\Delta}^{2}_{\text{Swin}} = \text{MLP}(\mathbf{\mu}_{\mathbf{X}},\mathbf{X})
\end{equation}
where $\tau^{2}_{\text{Swin}}\in\mathbb{R^{+}}$ is the scaling compensation and $\mathbf{\Delta}^{2}_{\text{Swin}}\in\mathbb{R}^{1\times d_k}$ is the shifting factor.

\begin{algorithm}[t!]
\caption{NSATP-1 (Swin)}
\label{algorithm_swin}
\KwIn{historical input length $N_{\text{p}}$, predicted output length $N_{\text{f}}$, model space $d_{\text{model}}$, number of 2D block $L_{\text{2D-block}}$, selected frequencies in FFT $k$, temporal feature $\mathbf{F}^{\text{temporal}}\in \mathbb{R}^{N_{\text{p}}\times C}$, contextual feature $\mathbf{F}^{\text{context}}\in \mathbb{R}^{N_{\text{p}}\times N_{\text{c}}}$, scheduled arrival time ${\mathbf{T}}^{\text{s}}\in\mathbb{R}^{N_{\text{f}}\times 1}$, the total number of windows $N_{\text{w}}$ with window size 3*3.}
\KwOut{arrival time $\hat{\mathbf{T}}^{\text{a}}$.}
$\mathbf{X}^{\prime}, \pmb{\mu}_{\mathbf{X}}, \pmb{\sigma}_{\mathbf{X}} = \text{Normalization}(\mathbf{F}^{\text{temporal}})$ as in Eqs.~\eqref{normal-3}-\eqref{normal-2}. 
\hfill $\triangleright$ $\mathbf{X}^{\prime}\in \mathbb{R}^{N_{\text{p}}\times C}$,  $\pmb{\mu}\in \mathbb{R}^{C\times 1}$, $\pmb{\sigma}\in \mathbb{R}^{C\times 1}$\\

Estimate $\tau^{1}_{\text{Swin}}$, $\mathbf{\Delta}^{1}_{\text{Swin}}$ by MLP as in Eqs.~\eqref{swin_factor_1}-\eqref{swin_factor_2}.
\hfill $\triangleright$ $\tau^{1}_{\text{Swin}}\in\mathbb{R^{+}}$, $\mathbf{\Delta}^{1}_{\text{Swin}}\in\mathbb{R}^{(3*3)\times d_{\text{model}}}$\\

Estimate $\tau^{2}_{\text{Swin}}$, $\mathbf{\Delta}^{2}_{\text{Swin}}$ by MLP as in Eqs.~\eqref{swin_ff_factor_1}-\eqref{swin_ff_factor_2}.
\hfill $\triangleright$ $\tau^{2}_{\text{Swin}}\in\mathbb{R^{+}}$, $\mathbf{\Delta}^{2}_{\text{Swin}}\in\mathbb{R}^{T\times d_{\text{model}}}$\\

$\hat{\mathbf{X}}_{\text{1D}}^{0} = \text{Embed}\left(\text{Concatenate}\left(\mathbf{X}^{\prime},\mathbf{F}^{\text{context}}\right)\right)$ by Conv1d-based feature embedding. 
\hfill $\triangleright$ $\mathbf{X}_{\text{1D}}^{0}\in\mathbb{R}^{T\times d_{\text{model}}}$, $T=N_{\text{p}}+N_{\text{f}}$\\

\For{$l \in \left\{1,2,..., L \right\}$}{
 1D$\rightarrow$2D: $(\hat{\mathbf{X}}_{\text{2D},1}^{l-1},...,\hat{\mathbf{X}}_{\text{2D},k}^{l-1}),(\mathbf{A}_{f_{1}}^{l-1},...,\mathbf{A}_{f_{k}}^{l-1})=\text{FFT}(\hat{\mathbf{X}}_{\text{1D}}^{l-1})$. 
\hfill $\triangleright$ $\mathbf{X}_{\text{2D},i}^{l-1} \in \mathbb{R}^{f_{i}\times p_{i}\times d_{\text{model}}}\ \ \forall i\in [1,k], i\in \mathbb{Z}$ \\
\For{$i \in \left\{1,2,..., k\right\}$}{
$\mathbf{X}_{\text{window},i}^{q,1}\leftarrow\text{Window partition}(\hat{\mathbf{X}}_{\text{2D},i}^{l-1})$
\hfill $\triangleright$ $\mathbf{X}_{\text{window},i}^{q,1} \in \mathbb{R}^{2\times 2\times d_{\text{model}}}\ \  \forall q\in[1,N_{\text{w}}], q\in \mathbb{Z}$\\
\For{$q\in [1,N_{\text{w}}]$}{
$\hat{\mathbf{Z}}_{\text{window},i}^{q,1} = \text{W-MSA}(\text{LN}(\mathbf{X}_{\text{window},i}^{q,1}), \tau^{1}_{\text{Swin}},\mathbf{\Delta}^{1}_{\text{Swin}})+ \mathbf{X}_{\text{window},i}^{q,1}$   \hfill $\triangleright$ self-attention inside window with NS\\
$\hat{\mathbf{X}}_{\text{window},i}^{q,1} = \text{MLP}((\text{LN}(\hat{\mathbf{Z}}_{\text{window},i}^{q,1})) + \hat{\mathbf{Z}}_{\text{window},i}^{q,1} $   \hfill $\triangleright$   feed forward 
}

$\hat{\mathbf{X}}_{\text{2D},i}^{l-1,1} \leftarrow \text{Combine}(\hat{\mathbf{X}}_{\text{window},i}^{q,1})$   \hfill $\triangleright$ $\hat{\mathbf{X}}_{\text{2D},i}^{l-1,1} \in \mathbb{R}^{f_{i}\times p_{i}\times d_{\text{model}}}$, $\hat{\mathbf{X}}_{\text{window},i}^{q,1} \in \mathbb{R}^{2\times 2\times d_{\text{model}}}\ \  \forall q\in[1,N_{\text{w}}], q\in \mathbb{Z}$\\

${\mathbf{X}}_{\text{window},i}^{q,2}\leftarrow \text{Partition}(\text{shifting}(\hat{\mathbf{X}}_{\text{window},i}^{q,1}))$  \hfill $\triangleright$ shifting the window and re-partition\\
\For{$q\in [1,N_{\text{w}}]$}{
$\hat{\mathbf{Z}}_{\text{window},i}^{q,2} = \text{SW-MSA}(\text{LN}(\mathbf{X}_{\text{window},i}^{q,2}), \tau^{1}_{\text{Swin}},\mathbf{\Delta}^{1}_{\text{Swin}})+ \mathbf{X}_{\text{window},i}^{q,2}$   \hfill $\triangleright$  self-attention after shifting window with NS\\
$\hat{\mathbf{X}}_{\text{window},i}^{q,2} = \text{MLP}((\text{LN}(\hat{\mathbf{Z}}_{\text{window},i}^{q,2})) + \hat{\mathbf{Z}}_{\text{window},i}^{q,2} $   \hfill $\triangleright$   feed forward 
}
$\hat{\mathbf{X}}_{\text{2D},i}^{l,2} \leftarrow \text{Combine}(\hat{\mathbf{X}}_{\text{window},i}^{q,2})$   \hfill $\triangleright$ $\hat{\mathbf{X}}_{\text{2D},i}^{l,2} \in \mathbb{R}^{f_{i}\times p_{i}\times d_{\text{model}}}$, $\hat{\mathbf{X}}_{\text{window},i}^{q,1} \in \mathbb{R}^{2\times 2\times d_{\text{model}}}\ \  \forall q\in[1,N_{\text{w}}], q\in \mathbb{Z}$\\

$\hat{\mathbf{X}}_{\text{1D},i}^{l}=\text{Reshape}(\hat{\mathbf{X}}_{\text{2D},i}^{l,2})+\hat{\mathbf{X}}_{\text{1D},i}^{l-1}$.
\hfill $\triangleright$ residual connection, $\mathbf{X}_{\text{1D},i}^{l}\in\mathbb{R}^{T\times d_{\text{model}}}$
}
$\hat{\mathbf{A}}_{\text{Top k}} =\text{Softmax}(\mathbf{A}_{\text{Top k}})$. \hfill $\triangleright$ rescale to $[0,1]$\\
$\hat{\mathbf{X}}_{\text{1D}}^{l} =\sum_{i=1}^{k} \hat{\mathbf{A}}_{f_{i}} \times \hat{\mathbf{X}}_{\text{1D}}^{l}$.
\hfill $\triangleright$  weighted sum, $\mathbf{X}_{\text{1D}}^{l}\in\mathbb{R}^{T\times d_{\text{model}}}$\\
}
Compensation $\mathbf{X}^{c}_{\text{Swin}} = \tau_{\text{Swin}}\hat{\mathbf{X}}_{\text{1D}}^{l}+\mathbf{\Delta}_{\text{Swin}}$ as in Eq.~\eqref{swin_compensation_2}. \hfill $\triangleright$  Non-stationary effect recovery\\

Calculate delays $\hat{\mathbf{T}}^{\text{d}}=\text{Trun}(\underset{d_{\text{model}}\rightarrow 1}{\text{Linear}}(\mathbf{X}^{c}_{\text{Swin}}))$.
\hfill $\triangleright$ truncation and linear projection back, $\hat{\mathbf{T}}^{\text{d}}\in\mathbb{R}^{N_{\text{f}}\times 1}$\\

Arrival time $\hat{\mathbf{T}}^{\text{a}}=\text{de-normalization}(\hat{\mathbf{T}}^{\text{d}})+{\mathbf{T}}^{\text{s}}$ as in Eq.~\eqref{multi_step}.
\hfill $\triangleright$  de-normalization and calculate ATP, $\hat{\mathbf{T}}^{\text{a}}\in\mathbb{R}^{N_{\text{f}}\times 1}$\\
\Return $\hat{\mathbf{T}}^{\text{a}}$
\end{algorithm}

\begin{figure}[t]
  \centering
  \includegraphics[width=1\textwidth]{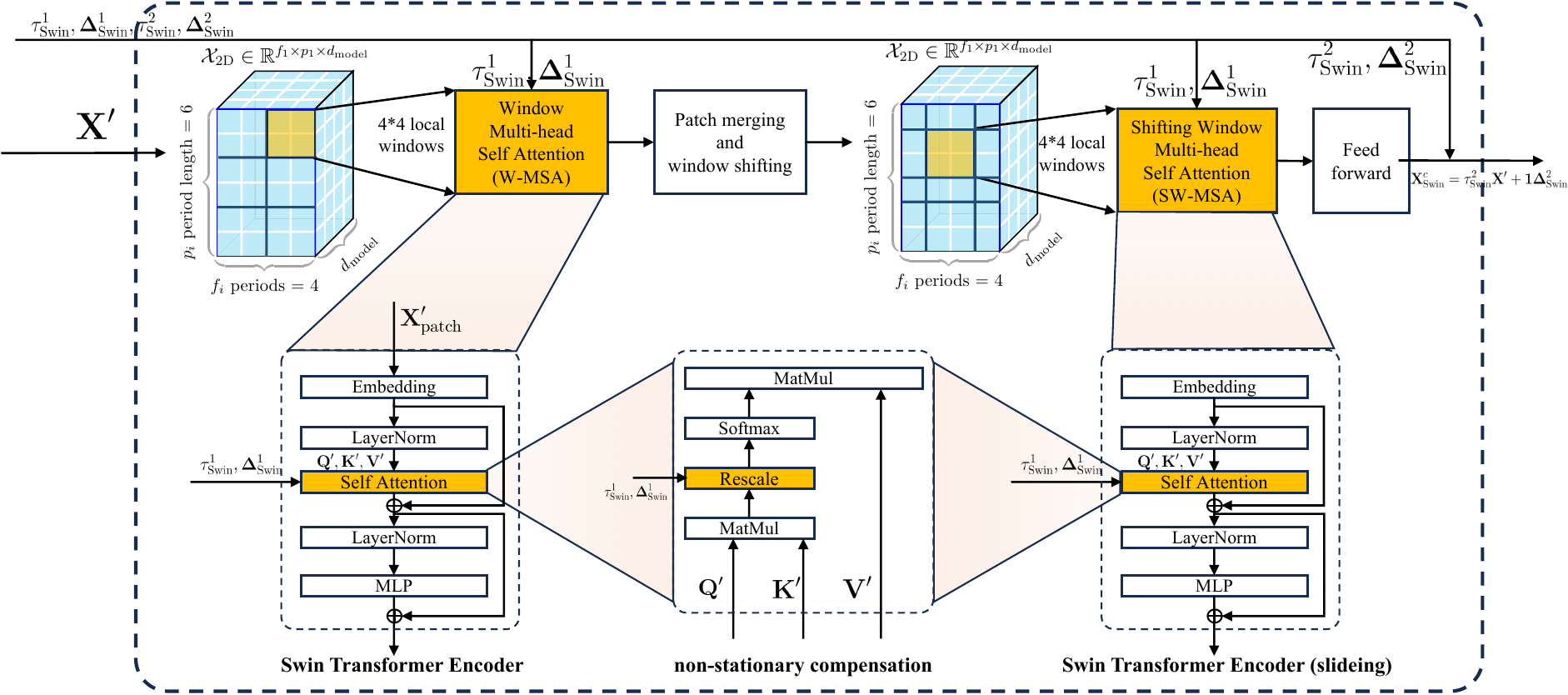}
  \caption{The Swin Transformer-based feature extraction for 2D image-like tensor with non-stationary effect recovery in Eq.~\eqref{swin_compensation_1} and Eq.~\eqref{swin_compensation_2}.}
  \label{fig_swin_illustration}
\end{figure}

\subsubsection{Compensation concatenation}
These two factors are applied on the output of Swin Transformer as follow:
\begin{equation}\label{swin_compensation_2}
    \mathbf{X}^{c}_{\text{Swin}} = \tau^{2}_{\text{Swin}}\mathbf{X}^{\prime}+ \mathbf{\Delta}^{2}_{\text{Swin}}
\end{equation}
where $\mathbf{X}^{c}_{\text{Swin}}$ is the tensor with over-stationarization compensation. The structure of \text{MLP} is similar to the compensation in Fig.~\ref{fig_MLP} but with different output size. Based on $\tau^{1}_{\text{Swin}}$, $\mathbf{\Delta}^{1}_{\text{Swin}}$, $\tau^{2}_{\text{Swin}}$ and $\mathbf{\Delta}^{2}_{\text{Swin}}$, the non-stationarity is recovered. Fig.~\ref{algorithm_swin} presents the Swin Transformer-based feature extraction for 2D image-like tensor with non-stationary recovery in Eq.~\eqref{swin_compensation_1} and Eq.~\eqref{swin_compensation_2}. The detailed procedure of NSATP-2 (Swin) is shown in Algorithm~\ref{algorithm_swin} and the relationship between the Swin Transformer-based feature extraction and non-stationary effect recovery is illustrated in Fig.~\ref{fig_swin_illustration}.

\section{Experiments}\label{experiments}
To validate the two proposed NSATP models, the operational data from the Dresden public transport system over a period of 125 days is collected. This section is divided into five parts: dataset, metrics, settings\&baseline, results, and discussion.

\subsection{Dataset}
The Dresden public transport system is operated by Dresdner Verkehrsbetriebe AG (DVB), which includes buses and trams. All vehicles send their locations to the operational system at intervals of approximately 15 seconds. In this work, raw data from the central system was collected over 125 days. The vehicle's delay at each stop can be calculated by matching the real-time operational data with the daily updated schedule information. Additionally, the details of the delay capture and the justifcation for input and parameter selection were presented in~\cite{liarrivalnet}.
\begin{figure}[ht!]
    \centering
    \begin{subfigure}[b]{0.5\textwidth}
        \centering
        \includegraphics[width=\textwidth]{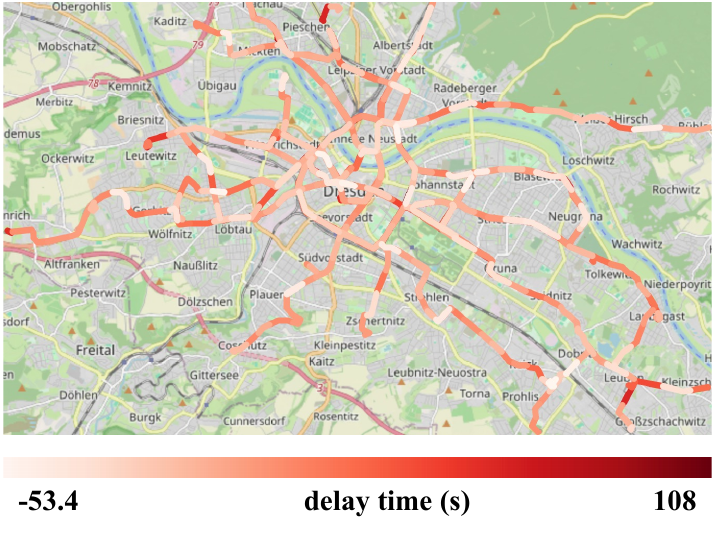}
        \caption{Link delay spatial distribution for tram.}
        \label{fig_delay_map_tram}
    \end{subfigure}
    
    \begin{subfigure}[b]{0.5\textwidth}
    \centering
    \includegraphics[width=\textwidth]{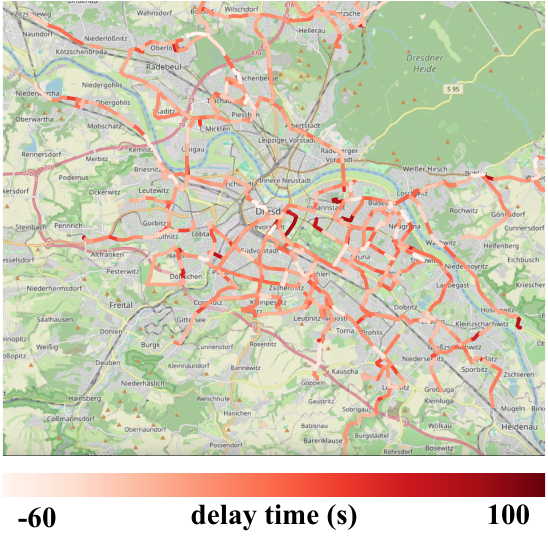}
        \caption{Link delay spatial distribution for bus.}
    \label{fig_delay_map_bus}
\end{subfigure}
    \caption{Average delays for trams and buses across all links.}
\end{figure}

\begin{figure}[ht]
    \centering
    \begin{subfigure}[b]{0.6\textwidth}
        \centering
        \includegraphics[width=\textwidth]{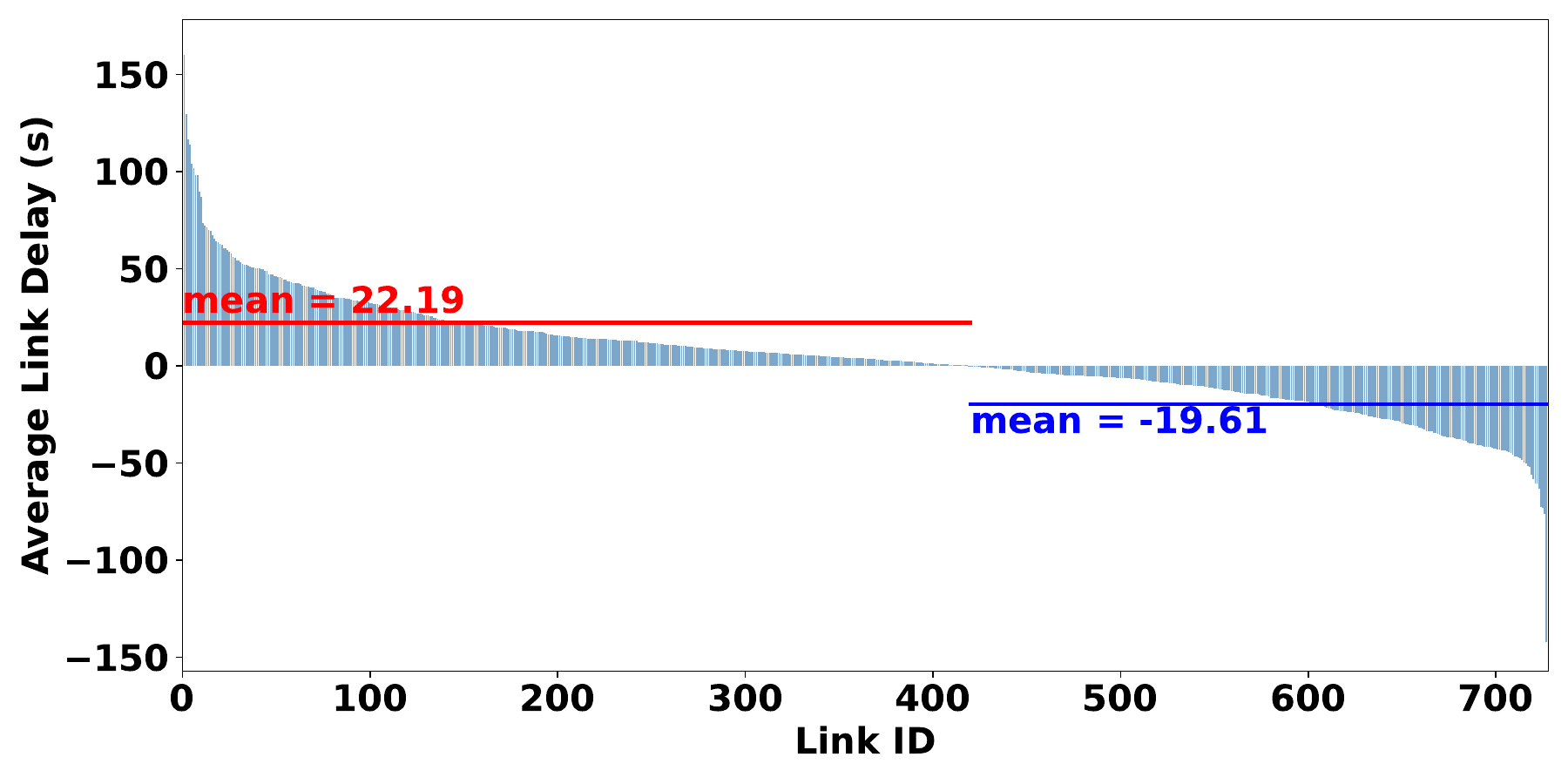}
        \caption{The rank of tram link delays.}
        \label{fig_delay_hist_tram}
    \end{subfigure}

    \begin{subfigure}[b]{0.6\textwidth}
        \centering
        \includegraphics[width=\textwidth]{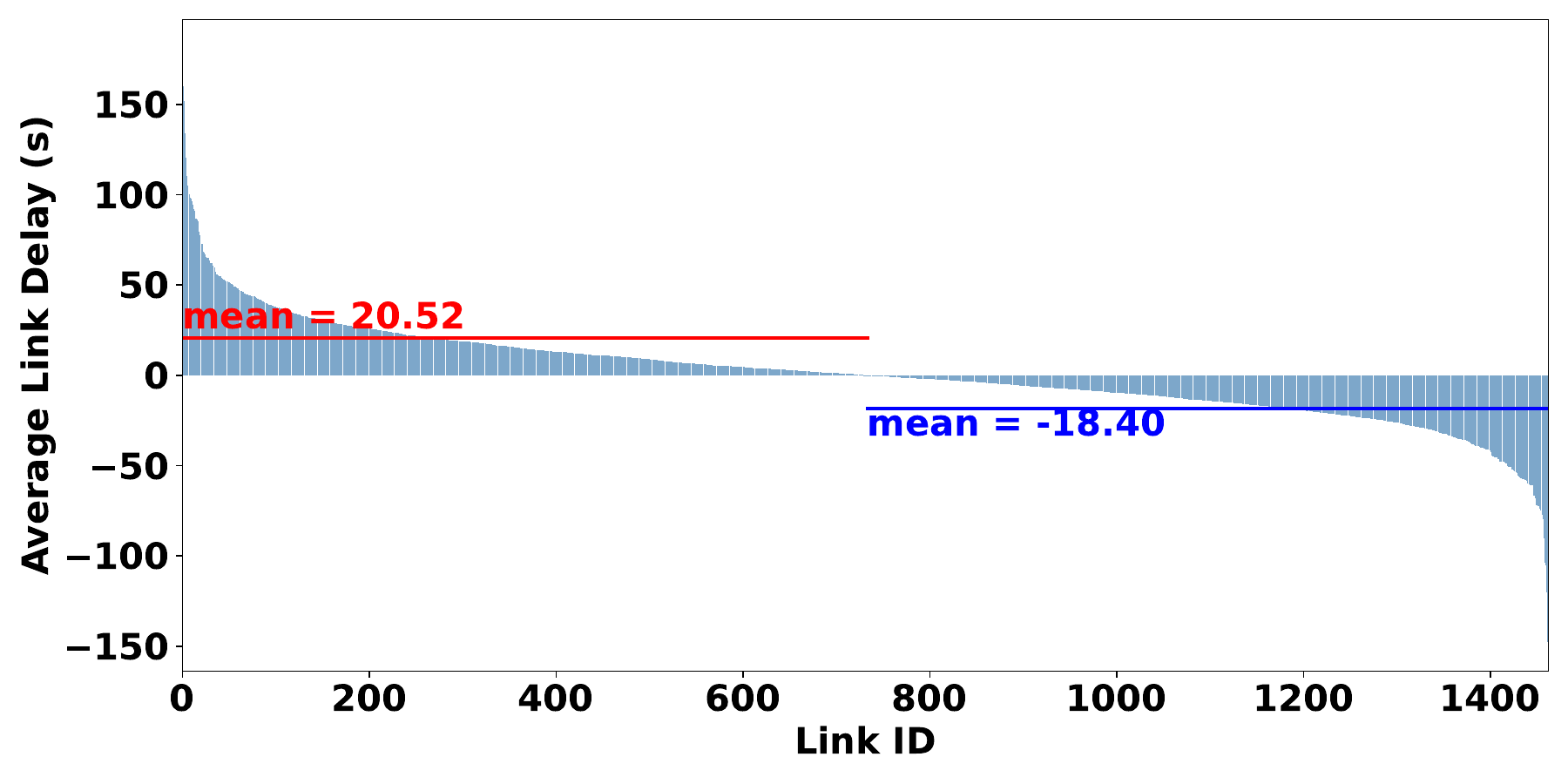}
        \caption{The rank of bus link delays.}
        \label{fig_delay_hist_bus}
    \end{subfigure}
    \caption{The rank of average link delays for tram and bus.}
\end{figure}

The processed dataset comprises $\sim$4.97M sequences. Fig.~\ref{fig_delay_map_tram} and Fig.~\ref{fig_delay_map_bus} illustrate the average delays for trams  and  buses across all links. From these two distributions of delay, it is evident that delays are commonly experienced by both trams and buses. In most of the links, the actual average travel time is not consistent with the scheduled time. Fig.~\ref{fig_delay_hist_tram} and Fig.~\ref{fig_delay_hist_bus} show the rank of average delays for tram and bus, which correspond to Fig.~\ref{fig_delay_map_tram} and Fig.~\ref{fig_delay_map_bus}. It indicate that delays vary within a range of approximately -100 to 150 seconds. The means of positive delays are 22.19 and 20.52 seconds for tram and bus, respectively, while the negative delays (early) are -19.61 and -18.40 seconds. Additionally, there are slightly more links with a positive average delay than those with a negative one. It underscores the importance of accurate public transport ATP in enhancing passenger experience.

\subsection{Metrics}
This work employs three metrics to evaluate the performance of multi-step ATP: root mean square error (RMSE), mean absolute error (MAE) and mean absolute percentage error (MAPE). Among these, RMSE is more sensitive to larger prediction errors compared to MAE. MAPE is used to evaluate the proportion of the prediction error relative to the actual values. The formulation of three matrics are shown as follows:
\begin{equation}
\text{RMSE} = \sqrt{\frac{1}{N_{\text{f}}}\sum_{t=i_{k}+1}^{i_{k}+N_{\text{f}}}  (\mathbf{T}^{\text{arrival}}_{i_{k}+t} - \hat{\mathbf{T}}^{\text{arrival}}_{i_{k}+t})^2}
\end{equation}
\begin{equation}
\text{MAE} = \frac{1}{N_{\text{f}}}\sum_{t=i_{k}+1}^{i_{k}+N_{\text{f}}} \left| \mathbf{T}^{\text{arrival}}_{i_{k}+t} - \hat{\mathbf{T}}^{\text{arrival}}_{i_{k}+t} \right|
\end{equation}
\begin{equation}
\text{MAPE} = \frac{100}{ N_{\text{f}}}\sum_{t=i_{k}+1}^{i_{k}+N_{\text{f}}} \left| \frac{\mathbf{T}^{\text{arrival}}_{i_{k}+t} - \hat{\mathbf{T}}^{\text{arrival}}_{i_{k}+t}}{\mathbf{T}^{\text{arrival}}_{i_{k}+t}} \right|
\end{equation}
where $N_{\text{f}}$ is number of future steps. $i_{k}$ is the stop index of $k^{th}$ sequence. $\hat{\mathbf{T}}^{\text{arrival}}_{i_{k}+t}$ and $\mathbf{T}^{\text{arrival}}_{i_{k}+t}$ are estimation and groundtruth values of $k^{th}$ testing sequences at $(i_{k}+t)^{th}$ stop, respectively.


\subsection{Settings and  baselines}
The parameters of basic arrival time predictor follow the experimental setting in~\cite{liarrivalnet}, which are detailed in Table~\ref{table_parameters}. As for the MLP-based compensation, it consists of two 128-dimensional hidden layers.
\begin{table}[htbp]
\caption{Experimental Settings}
\label{table_parameters}
\centering
\begin{tabularx}{0.8\linewidth}{lXX}
\toprule
\textbf{Description} & \textbf{Notation} & \textbf{Value} \\
\midrule
learning rate & - & 0.001 \\
embedding feature length & $d_{\text{model}}$ &16 \\
number of 2d block & - & 2 \\
selected frequency & $k$ & 3 \\
number of kernels in CNN & - & 6 \\
input series length & $N_{\text{p}}$ & 10 \\
output series length & $N_{\text{f}}$ & 5, 10 \\
local window size & - & 2 \\
minimum delay& - & -300 \\
maximum  delay& - & 1000\\
loss function& - & MSE\\
batch size& - & 256\\
number of MLP layers & - & 2\\
MLP hidden dimensions & - & 128\\
threshold in the signal selection (m)& - & 20 \\
\bottomrule
\end{tabularx}
\end{table}

To reflect the outstanding performance of the proposed NSATP, several methods are selected as baselines, including smoothing-based method, RNN-based method, attention-based method, basic predictor \textit{ArrivalNet} and learnable normalization-based method, which  are listed as follows:
\begin{itemize}
\item \textbf{Auto-regressive integrated moving average (ARIMA)}. It is a standard time series forecasting method that incorporates smoothing techniques. ARIMA combines autoregression with moving averages, which enables the establishment of a relationship between historical observations and future predictions~\cite{box1970distribution}.
\item \textbf{LSTM}. As a variant of the RNN model, it can model both long-term and short-term memories by the forget gate~\cite{hochreiter1997long}. The LSTM model and its derivatives have been widely used in predicting public transport arrival times.
\item \textbf{Vanilla Transformer}. It is developed based on the attention mechanism that can effectively capture the relationships between features at non-adjacent time instants in long sequences~\cite{vaswani2017attention}. It is widely applied to a variety of tasks, including natural language processing (NLP) and text translation~\cite{devlin2018bert}. In this work, the plain Vanilla Transformer in~\cite{vaswani2017attention} is applied. 
\item \textbf{Temporal convolutional network (TCN)}. It is a CNN-based time series prediction method. TCN extracts the useful information by applying convolution operations along the temporal dimension, which has been successfully applied in sequential action recognition and trajectory prediction~\cite{bai2018empirical}.
\item \textbf{\textit{ArrivalNet-1} (CNN)}. It is the basic multi-step ATP with CNN vision backbone. 
\item \textbf{\textit{ArrivalNet-2} (Swin)}. It is the basic multi-step ATP with Swin Transformer vision backbone. 
\item \textbf{RevIN}. It applies instance normalization with learnable affine parameters to each input and restores the statistics to the corresponding output, which makes each series follow a similar distribution~\cite{kim2021reversible}.
\end{itemize}

As for the proposed NSATP, two variants are presented:
\begin{itemize}
\item \textbf{NSATP-1 (CNN)}. It is the proposed non-stationary arrival time predictor, which is developed based on \textit{ArrivalNet-1} (CNN) and equipped with the compensation module designed in Section~\ref{compensation_cnn}.
\item \textbf{NSATP-2 (Swin)}. It is the proposed non-stationary arrival time predictor, which is developed based on \textit{ArrivalNet-2} (Swin) and equipped with the compensation module designed in Section~\ref{compensation_swin}.
\end{itemize}
\subsection{Results}
This work focuses on the impact of stationarity on time series. The augmented dickey-fuller (ADF) stationarity indices for all sequences of buses and trams are calculated, where a smaller ADF value indicates higher stationarity. The formulation of ADF is detailed in Appendix~\ref{appendix_adf}. Fig.~\ref{fig_adf} presents ADF values before and after the series stationarization when the sequence length is 20.  With the normalization, the average ADF value of tram decreases from -3.20 to -7.95 (148\%), while that of bus decrease from -2.37 to -5.20 (119\%). It suggests that the series stationarization will increase the stationarity of sequence, which may lead to the over-stationarization issue in the predictor. 

\begin{table*}[t!]
  \centering
  \caption{Comparative Results of Tram Arrival Time Prediction (lower is better).}
  \label{table_tram}
  \begin{tabular}{
  l>{\centering\arraybackslash}p{1.4cm}
  >{\centering\arraybackslash}p{1.4cm}
  >{\centering\arraybackslash}p{1.4cm}
  >{\centering\arraybackslash}p{1.4cm}
  >{\centering\arraybackslash}p{1.4cm}
  >{\centering\arraybackslash}p{1.4cm}}
    \toprule
    \multirow{2}{*}{Method} &
      \multicolumn{3}{c}{Length 10$\rightarrow$5} &
      \multicolumn{3}{c}{Length 10$\rightarrow$10} \\ 
      \cmidrule(lr){2-4}\cmidrule(lr){5-7}
      & RMSE (s) & MAE (s) & MAPE (\%) 
      & RMSE (s) & MAE (s) & MAPE (\%) \\
    \midrule
    ARIMA                 & 64.3 & 55.2 & 4.94 & 79.3 & 61.2 & 6.71 \\
    LSTM                  & 60.4 & 49.9 & 4.86 & 72.4 & 57.7 & 5.12 \\
    Transformer           & 49.8 & 38.8 & 3.83 & 69.1 & 50.7 & 4.32 \\
    TCN                   & 49.2 & 42.5 & 3.59 & 68.5 & 48.6 & 3.93 \\
    RevIN                 & 46.1 & 32.4 & 2.47 & 58.2 & 37.9 & 2.64 \\
    \textit{ArrivalNet} (CNN)  & 46.2 & \underline{31.5} & \underline{2.39} & 57.4 & \underline{37.4} & \underline{2.55} \\
    \textit{ArrivalNet} (Swin) & 46.4 & 32.7 & 2.54 & 56.8 & 38.6 & 2.66 \\
    NSATP-1 (CNN)         & \textbf{45.1} & \textbf{30.7} & \textbf{2.30} & \underline{55.8} & \textbf{36.4} & \textbf{2.47} \\
    NSATP-2 (Swin)        & \underline{45.3} & 32.3 & 2.48 & \textbf{54.9} & 37.5 & 2.60 \\
    \bottomrule
  \end{tabular}
  \vspace{0.5ex}
  \begin{tablenotes}
    \item The $1^{\text{st}}$/$2^{\text{nd}}$ best results are in \textbf{bold}/\underline{underline}.
  \end{tablenotes}
\end{table*}

\begin{figure}[t]
  \centering
  \includegraphics[width=0.7\textwidth]{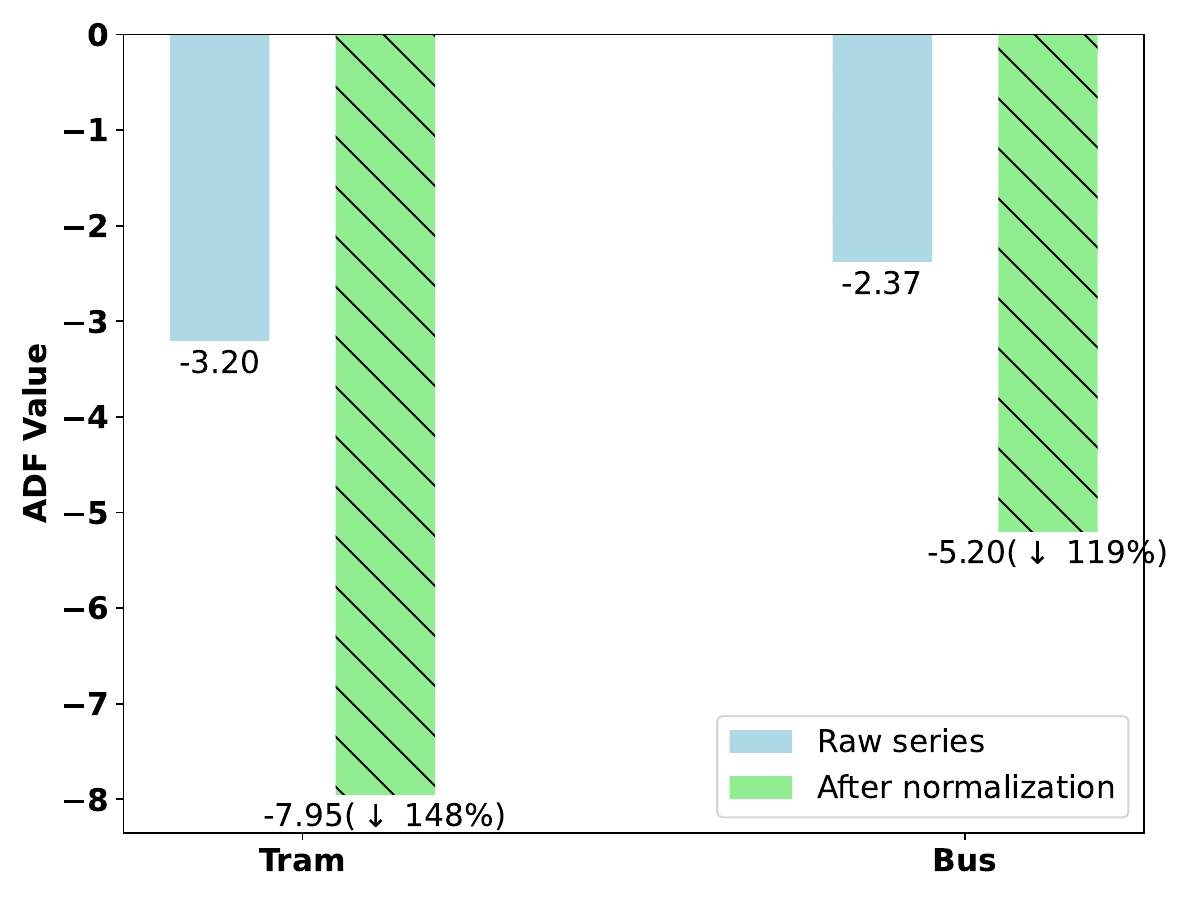}
  \caption{The augmented dickey-fuller (ADF) of raw and normalized series.}
  \label{fig_adf}
\end{figure}

Table~\ref{table_tram} presents the comparative results of all methods for tram ATP, respectively. ARIMA obtains the worst prediction, which suggests the averaging smoothing solution is not fit for tram ATP compared to deep learning-based algorithms. \textit{ArrivalNet}-based and NSATP-based approaches achieve all $1^{st}$ best and $2^{nd}$ best performance in RMSE, MAE and MAPE. It indicates that the basic model outperforms the smoothing-based and one-dimensional deep learning-based methods. Compared to LSTM, Vanilla Transformer and TCN in the average of two lengths (10$\rightarrow$5 and 10$\rightarrow$10), the two-dimensional based methods (\textit{ArrivalNet} (Swin), \textit{ArrivalNet} (CNN), NSATP-1 (CNN), NSATP-2 (Swin)) at least decrease the performance of RMSE by 14.87\%, MAE by 25.03\% and MAPE by 36.44\%, respectively. Comparing \textit{ArrivalNet} and NSATP, the NSATP-1 (CNN) obtains the best results in all three metrics. Except for the RMSE of length 10$\rightarrow$10, the performance of CNN-based \textit{ArrivalNet} and NSATP is better than Swin Transformer-based variants, which shows that 2D CNN is good at extracting useful information from the 2D image-like features. This finding applies to architectures with and without non-stationarity effect recovery. The performance of RevIN is basically same as  that of \textit{ArrivalNet}, which indicates that the normalization without learnable parameters (series stationarization) is sufficient to stationarize time series. Table~\ref{table_bus} shows the performance of bus ATP. Overall, it reflects a similar trend in comparative results. The proposed NSATP-1 (CNN) and NSATP-2 (Swin) are better than the corresponding basic models. The quantitative results in Table~\ref{table_tram} and Table~\ref{table_bus} illustrate the superiority of non-stationarity effect recovery.

\begin{table*}[ht]
  \centering
  \caption{Comparative Results of Bus Arrival Time Prediction (lower is better).}
  \label{table_bus}
  \begin{tabular}{
  l>{\centering\arraybackslash}p{1.4cm}
  >{\centering\arraybackslash}p{1.4cm}
  >{\centering\arraybackslash}p{1.4cm}
  >{\centering\arraybackslash}p{1.4cm}
  >{\centering\arraybackslash}p{1.4cm}
  >{\centering\arraybackslash}p{1.4cm}}
    \toprule
    \multirow{2}{*}{Method} &
      \multicolumn{3}{c}{Length 10$\rightarrow$5} &
      \multicolumn{3}{c}{Length 10$\rightarrow$10} \\ 
      \cmidrule(lr){2-4}\cmidrule(lr){5-7}
      & RMSE (s) & MAE (s) & MAPE (\%) 
      & RMSE (s) & MAE (s) & MAPE (\%) \\
    \midrule
    ARIMA                 & 66.1 & 53.8 & 5.84 & 83.3 & 61.4 & 8.73 \\
    LSTM                  & 63.7 & 49.5 & 5.32 & 74.2 & 59.3 & 6.30 \\
    Transformer           & 53.1 & 39.2 & 3.94 & 65.1 & 51.0 & 4.27 \\
    TCN                   & 52.8 & 41.4 & 3.91 & 67.9 & 49.8 & 4.39 \\
    RevIN                 & 48.6 & 34.1 & 3.43 & 57.7 & 38.1 & 3.72 \\
    \textit{ArrivalNet} (CNN)  & 48.3 & \underline{33.4} & \underline{3.35} & 58.3 & 38.3 & 3.74 \\
    \textit{ArrivalNet} (Swin) & 48.7 & 34.2 & 3.42 & \underline{57.3} & 38.5 & 3.65 \\
    NSATP-1 (CNN)         & \textbf{47.1} & \textbf{33.2} & \textbf{3.29} & \underline{57.3} & \underline{37.5} & \underline{3.64} \\
    NSATP-2 (Swin)        & \underline{47.4} & 34.3 & 3.38 & \textbf{56.2} & \textbf{37.3} & \textbf{3.58} \\
    \bottomrule
  \end{tabular}
  \vspace{0.5ex}
  \begin{tablenotes}
    \item The $1^{\text{st}}$/$2^{\text{nd}}$ best results are in \textbf{bold}/\underline{underline}.
  \end{tablenotes}
\end{table*}


\begin{table}[h!]
  \begin{center}
    \caption{The corresponding influence of non-stationary recovery approach in tram ATP. (The decrease and increase of prediction error are indicated in $\downarrow$ and $\uparrow$, respectively.)}
    \label{table_decrease_tram}
    \begin{tabular}{cccc} 
    \toprule
      \textbf{Metric}  & \textbf{Length}  & \textbf{NSATP-1 (CNN)}& \textbf{NSATP-2 (Swin)}\\
      \midrule
      \multirow{2}{*}{\parbox{1.5cm}{\centering RMSE}}
 & 10$\rightarrow$5   & $\downarrow$ 2.38\% & $\downarrow$ 2.37\%\\ 
      
 & 10$\rightarrow$10    &$\downarrow$ 2.79\% & $\downarrow$ 3.35\% \\
      \hline
      \multirow{2}{*}{\parbox{1.5cm}{\centering MAE}} 
      & 10$\rightarrow$5    & $\downarrow$ 2.54\%  &$\downarrow$ 1.22\%  \\ 
       & 10$\rightarrow$10   &$\downarrow$ 2.68\% & $\downarrow$ 2.85\% \\
      \hline
      \multirow{2}{*}{\parbox{1.5cm}{\centering MAPE}}    
      & 10$\rightarrow$5   & $\downarrow$ 3.77\% &$\downarrow$ 2.36\%  \\ 
        & 10$\rightarrow$10  & $\downarrow$ 3.14\%&$\downarrow$ 2.26\%  \\
      \bottomrule
    \end{tabular}
  \end{center}
\end{table}

\begin{table}[h!]
  \begin{center}
    \caption{The Corresponding Influence of Non-stationary recovery approaches in Bus ATP. (The decrease and increase of prediction error are indicated in $\downarrow$ and $\uparrow$, respectively.)}
    \label{table_decrease_bus}
    \begin{tabular}{cccc} 
    \toprule
      \textbf{Metric}  & \textbf{Length}  & \textbf{NSATP-2 (CNN)}& \textbf{NSATP-1 (Swin)}\\
      \midrule
      \multirow{2}{*}{\parbox{1.5cm}{\centering RMSE}}
 & 10$\rightarrow$5  & $\downarrow$ 2.48\% & $\downarrow$ 2.67\% \\ 
      
 & 10$\rightarrow$10   &$\downarrow$ 1.72\%  & $\downarrow$ 1.92\% \\
      \hline
      \multirow{2}{*}{\parbox{1.5cm}{\centering MAE}} 
      & 10$\rightarrow$5      & $\downarrow$ 0.60\% &$\uparrow$ 0.29\% \\ 
       & 10$\rightarrow$10  &$\downarrow$ 2.09\%  & $\downarrow$ 3.12\% \\
      \hline
      \multirow{2}{*}{\parbox{1.5cm}{\centering MAPE}}    
      & 10$\rightarrow$5    & $\downarrow$ 1.79\% &$\downarrow$ 1.17\% \\ 
        & 10$\rightarrow$10   & $\downarrow$ 2.67\%&$\downarrow$ 1.92\%\\
      \bottomrule
    \end{tabular}
  \end{center}
\end{table}

\begin{figure}[h!]
  \centering
  \begin{subfigure}[b]{0.48\textwidth}
    \includegraphics[width=1\textwidth]{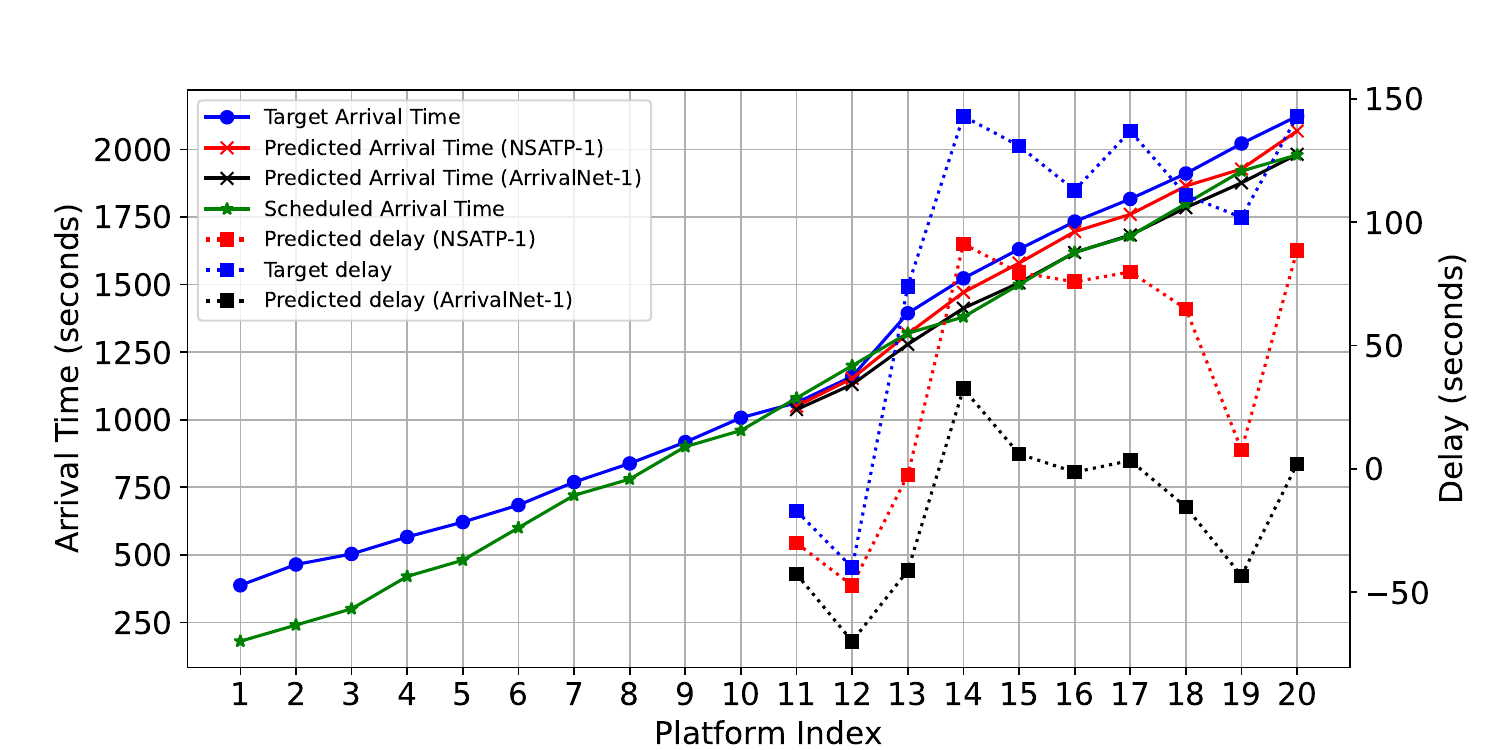}
    \caption{Tram ATP: Case 1.}
    \label{tram_case_1}
  \end{subfigure}
  \begin{subfigure}[b]{0.48\textwidth}
    \includegraphics[width=1\textwidth]{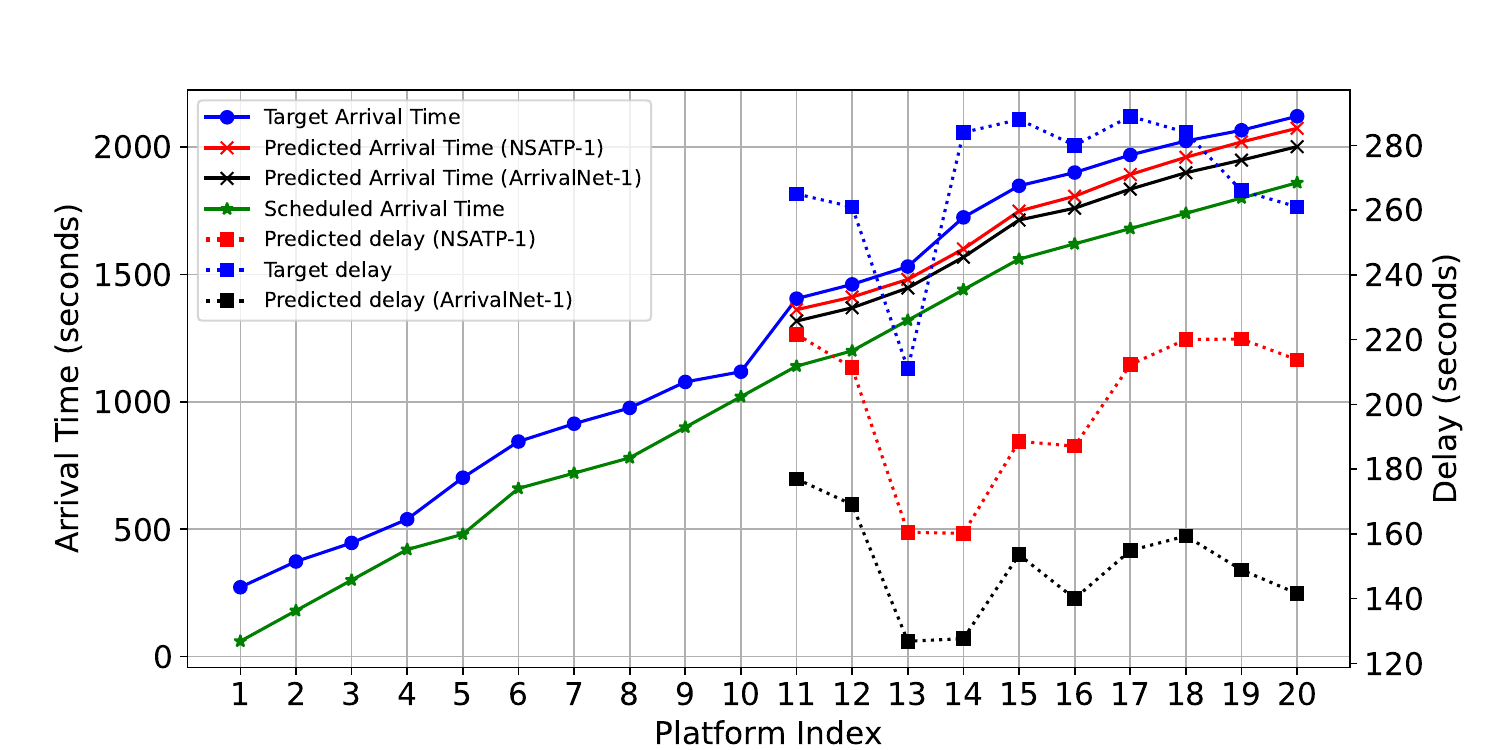}
    \caption{Tram ATP: Case 2.}
    \label{tram_case_2}
  \end{subfigure}

  \begin{subfigure}[b]{0.48\textwidth}
    \includegraphics[width=1\textwidth]{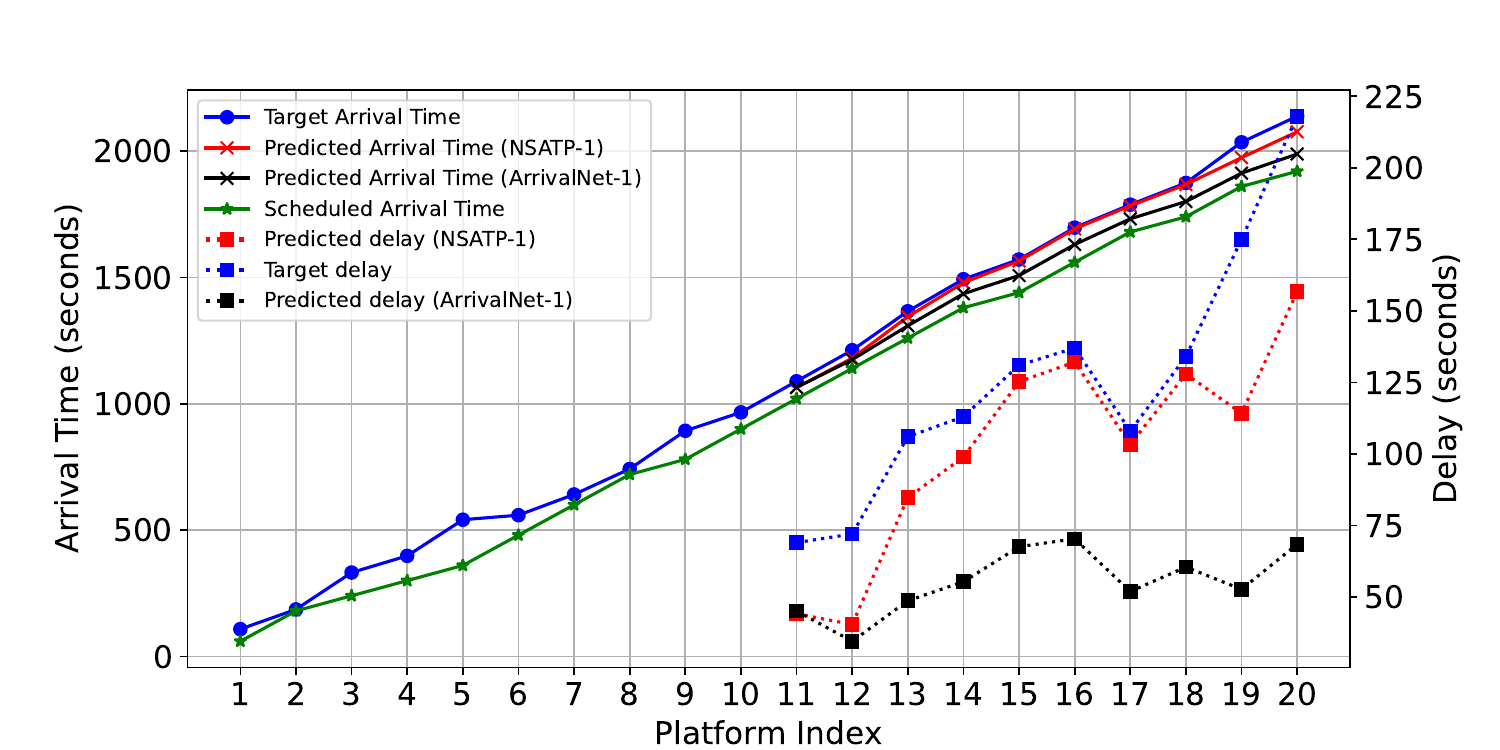}
    \caption{Tram ATP: Case 3.}
    \label{tram_case_3}
  \end{subfigure}
    \begin{subfigure}[b]{0.48\textwidth}
    \includegraphics[width=1\textwidth]{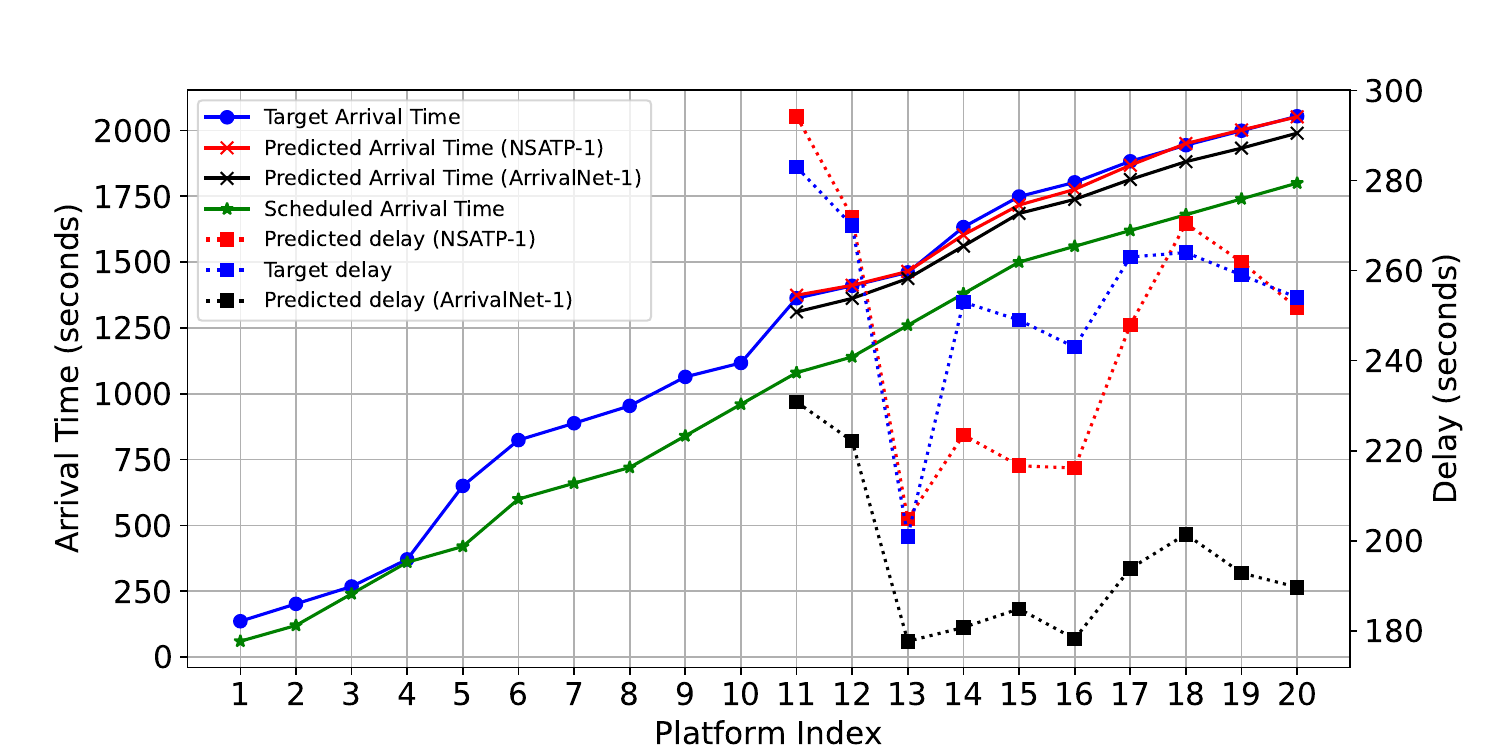}
    \caption{Tram ATP: Case 4.}
    \label{tram_case_4}
  \end{subfigure}
  \caption{The case study of tram ATP (model: NSATP-1, length: 10$\rightarrow$10).}
  \label{tram_casestudy}
\end{figure}

\begin{figure}[h!]
  \centering
  \begin{subfigure}[b]{0.48\textwidth}
    \includegraphics[width=1\textwidth]{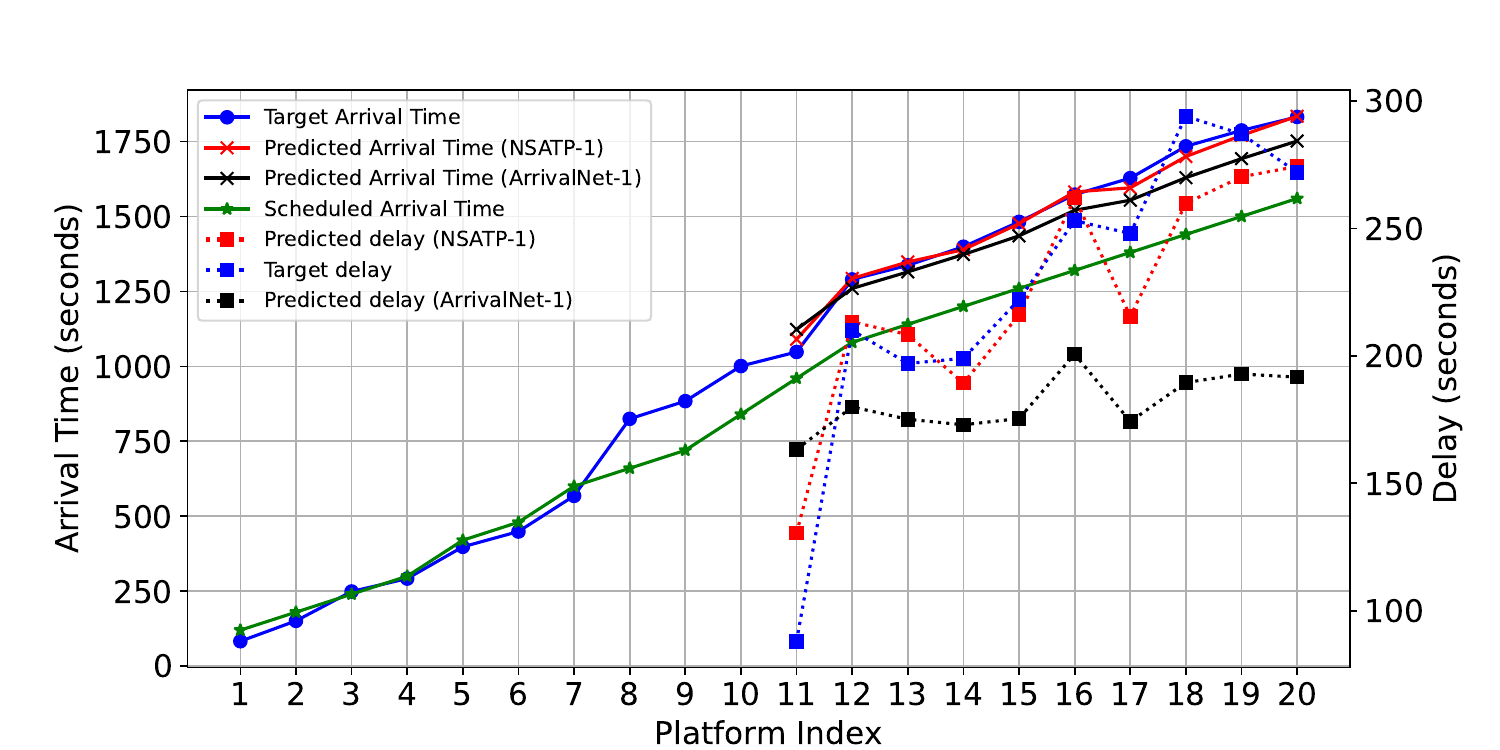}
    \caption{Bus ATP: Case 1.}
    \label{bus_case_1}
  \end{subfigure}
  \begin{subfigure}[b]{0.48\textwidth}
    \includegraphics[width=1\textwidth]{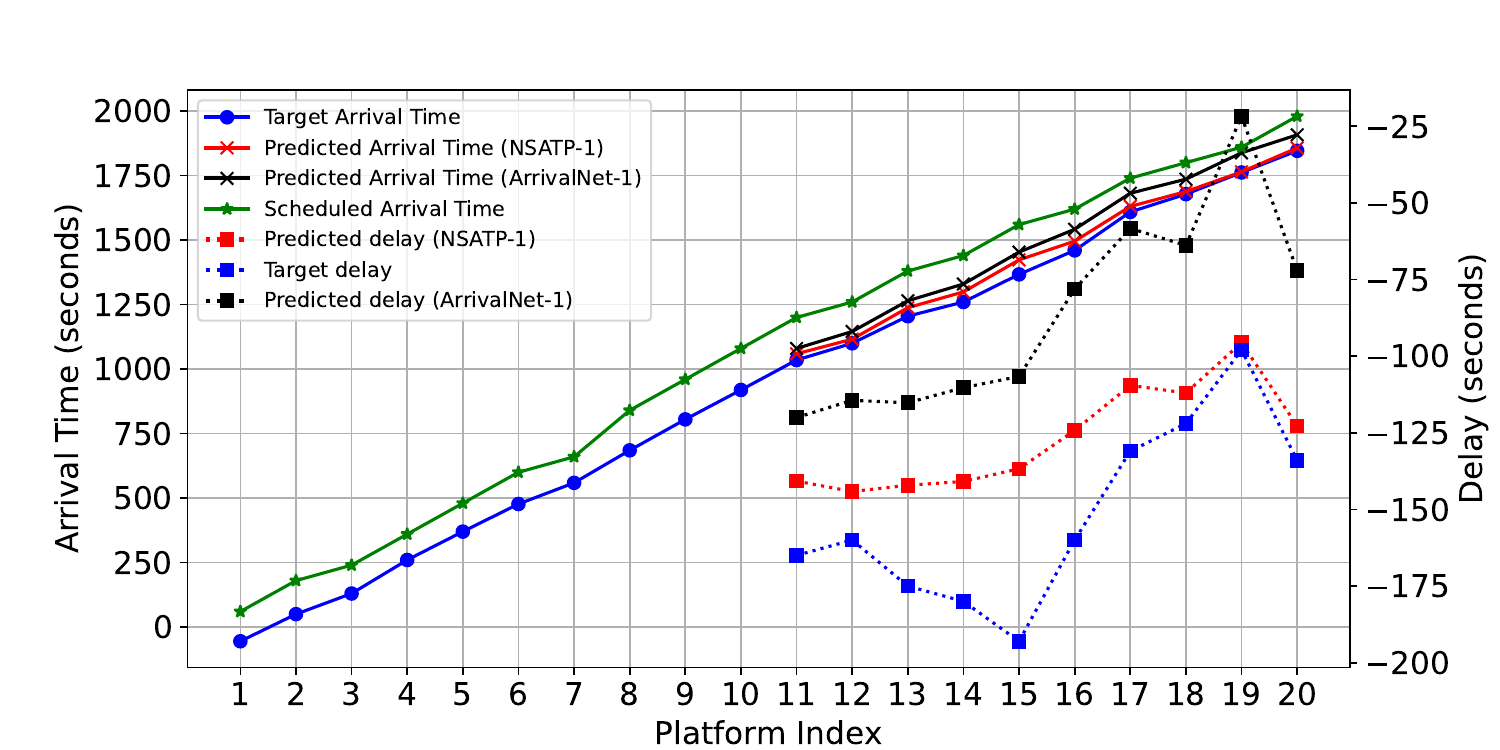}
    \caption{Bus ATP: Case 2.}
    \label{bus_case_2}
  \end{subfigure}

  \begin{subfigure}[b]{0.48\textwidth}
    \includegraphics[width=1\textwidth]{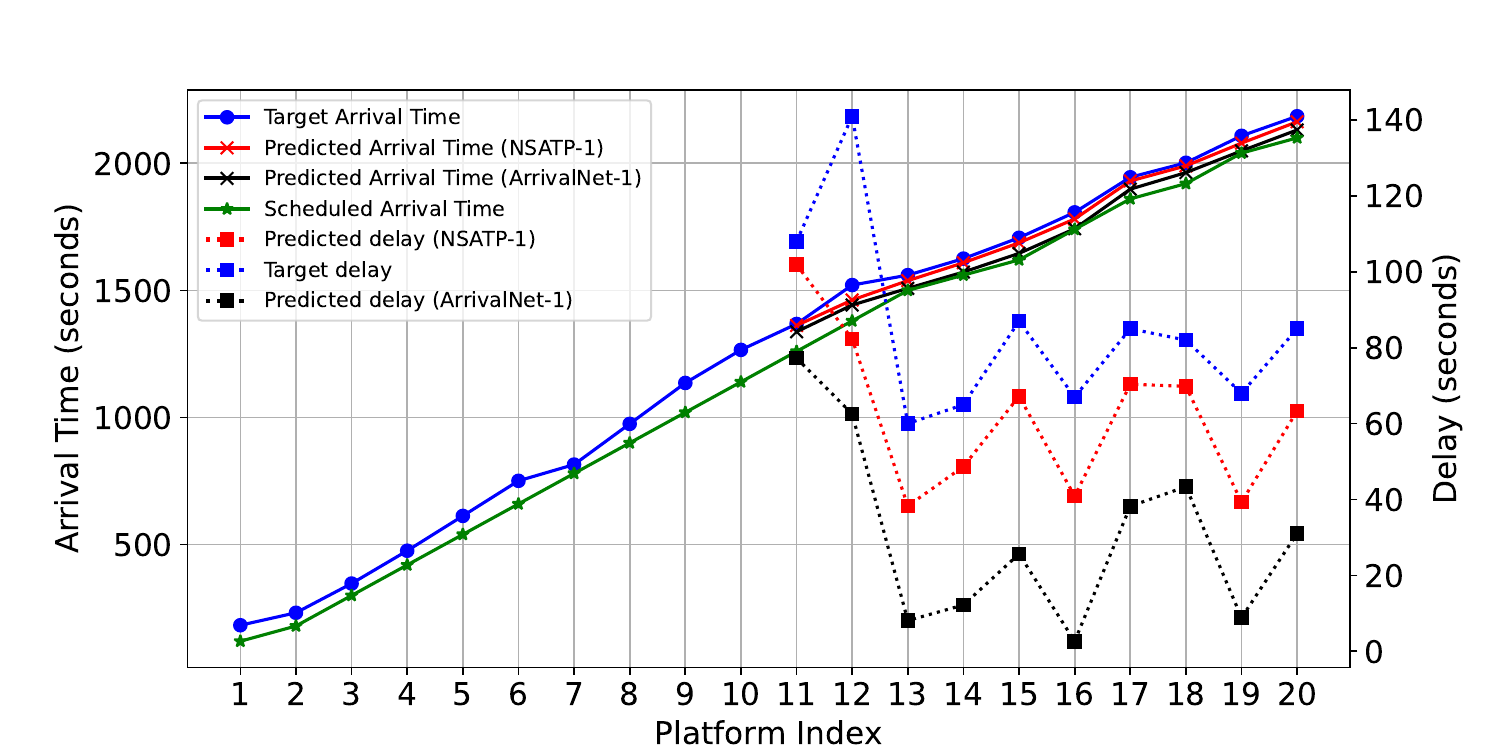}
    \caption{Bus ATP: Case 3.}
    \label{bus_case_3}
  \end{subfigure}
  \begin{subfigure}[b]{0.48\textwidth}
    \includegraphics[width=1\textwidth]{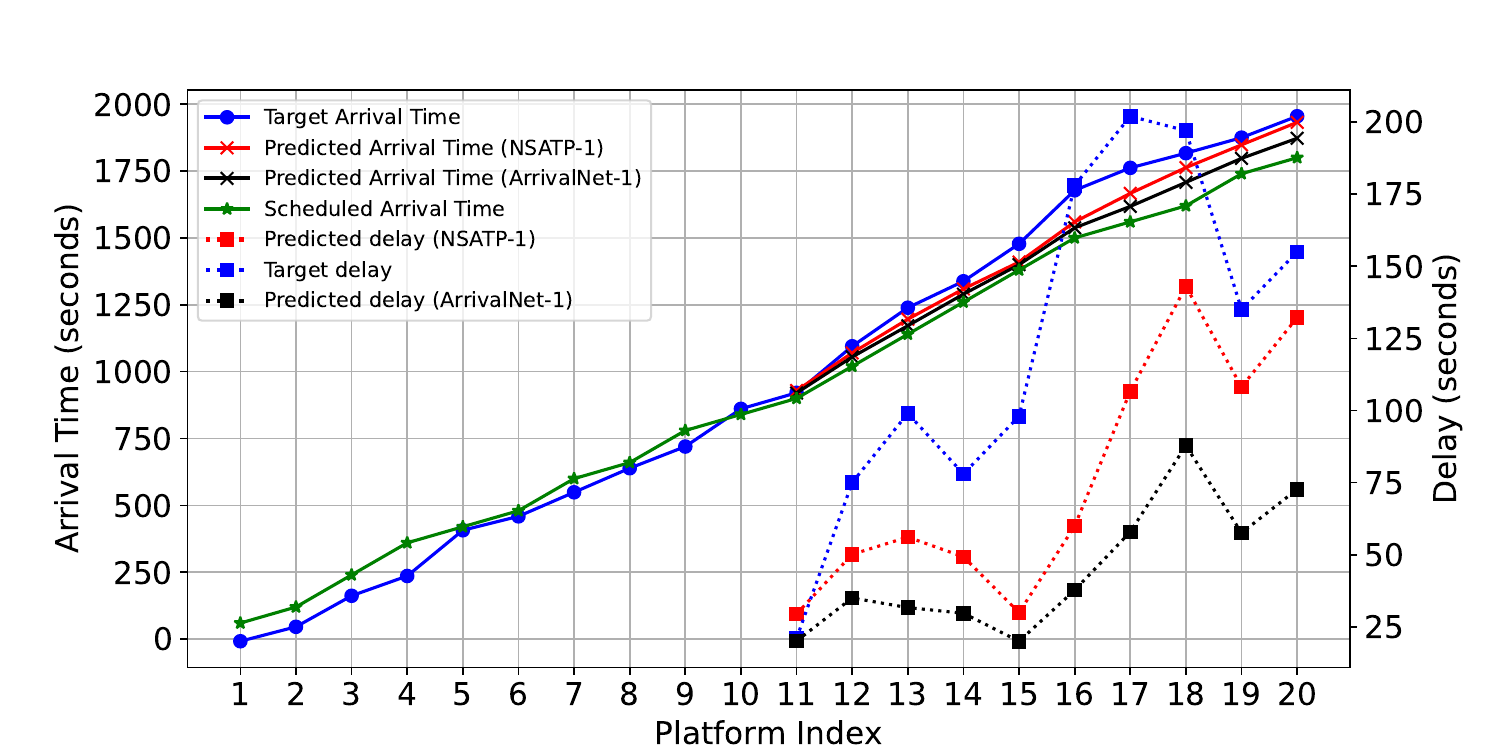}
    \caption{Bus ATP: Case 4.}
    \label{bus_case_4}
  \end{subfigure}
  \caption{The case study of bus ATP (model: NSATP-1, length: 10$\rightarrow$10).}
    \label{bus_casestudy}
\end{figure}

\subsection{Discussion}
\subsubsection{The positive improvement of non-stationary effect}
As for the positive effect of the over-stationarization compensation, compared to \textit{ArrivalNet-1} and \textit{ArrivalNet-2}, the corresponding improvements of NSATP-1 and NASTP-2 in three metrics are detailed in Table~\ref{table_decrease_tram}. Both Swin-based and CNN-based NSATP achieve a decrease in the prediction error compared to basic models. It demonstrates the proposed compensatory approaches play a positive role in coping with the over-stationarization issue. In detail, the NSATP model decreases by at least 2.37\%, 1.22\% and 2.26\% in RMSE, MAE and MAPE, respectively. Table~\ref{table_decrease_bus} presents similar comparative results of bus ATP. Overall, compared to tram ATP, the improvements in bus ATP are smaller. It decreases by at least 1.72\%, 0.60\% and 1.17\% in RMSE, MAE and MAPE, respectively. When the prediction length is 10$\rightarrow$5 and the metric is MAE, the proposed NSATP-1 (Swin) increases the error, possibly due to the combined effect of multiple compensatory mechanisms. 

\subsubsection{Case study of bus and tram ATP}
To demonstrate the performance of the proposed NSATP in specific scenarios, Fig.~\ref{tram_casestudy} and Fig.~\ref{bus_casestudy} show some cases for tram and bus, respectively. The lengths of historical input and predicted sequence are both 10. In each case, the ground truth arrival time, scheduled arrival time, predicted arrival time, ground truth delay and predicted delay are presented in detail. The figures show that for buses and trams, the ground truth arrival times and scheduled arrival times follow similar trends, but do not exactly match values. It underscores the necessity of multi-step ATP. Comparing \textit{ArrivalNet-1} (CNN) and NSATP-1 (CNN), both methods generate similar trends. However, the predictions from NSATP-1 are closer to the ground truth. In most cases, there is a relatively constant bias between the predictions of \textit{ArrivalNet-1} (CNN) and NSATP-1 (CNN), which can be attributed to the shifting module in the over-stationarization compensation.

\subsubsection{The relative stationarity of different methods in two datasets (bus and tram)}
\begin{figure}[t]
  \centering
  \includegraphics[width=0.7\textwidth]{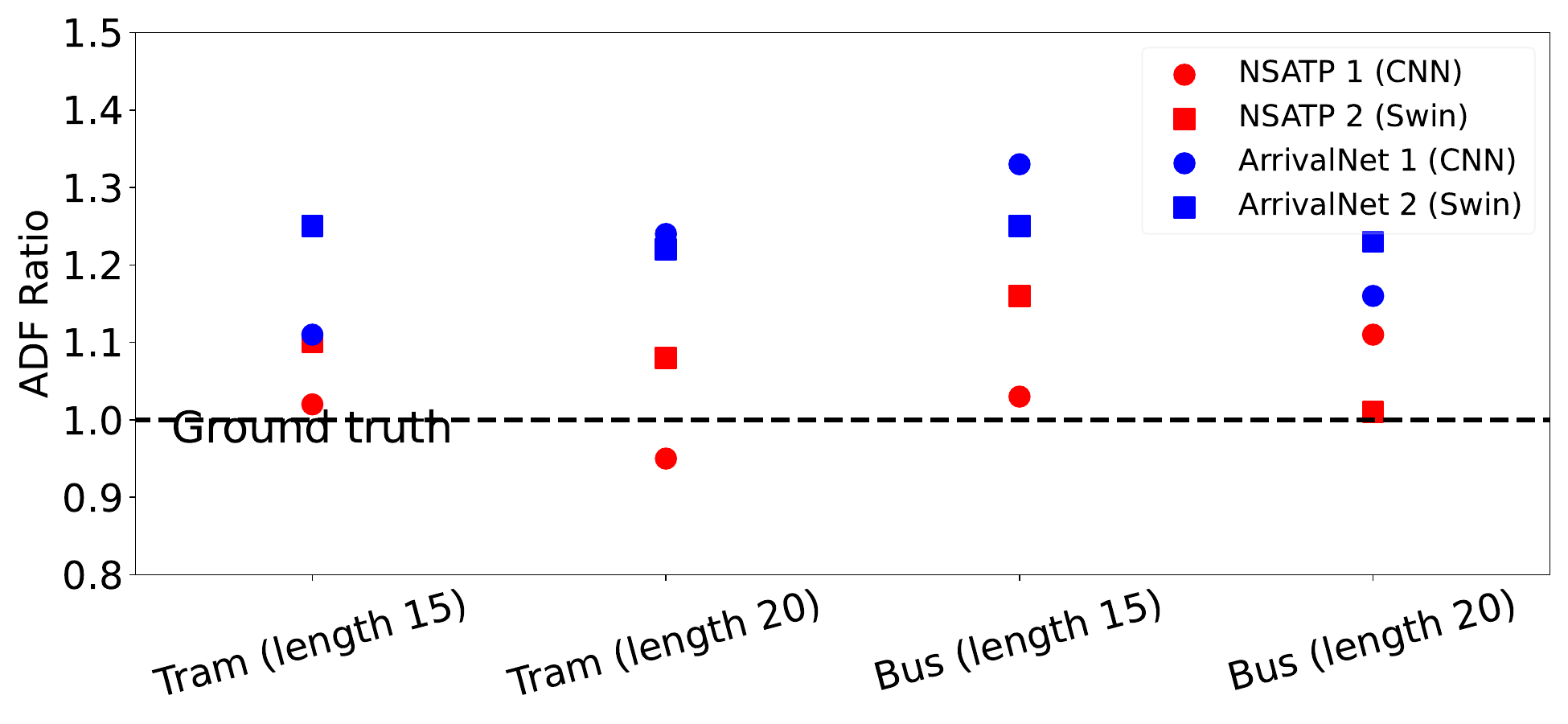}
  \caption{The ADF ratio of prediction and ground truth.}
  \label{fig_ns_ratio}
\end{figure}

To validate the influence of over-stationarization compensation on bus/tram ATP from the statistical view, Fig.~\ref{fig_ns_ratio} shows the ratio of ADF with different lengths and methods. The ADF ratio is the relative stationarity of the full sequence with predicted values in the future step and the full ground truth sequence in the same length. A larger ratio indicates that the model tends to predict ATP with higher stability, whereas a smaller ratio suggests that the model tends to increase the non-stationarity of the sequence. In Fig.~\ref{fig_ns_ratio}, the results from two ArrivalNet-based variants exceed 1 ( from 1.11 to 1.25), which suggests the models without  compensation generate sequential outputs with a higher stationarity. The results of NSATP -1 (CNN) in 4 experiments are closer to 1 (from 0.95 to 1.16). It indicates that the compensation in Section~\ref{compensation_cnn} and Section~\ref{compensation_swin} can extract useful non-stationary information from the raw sequence. In comparison to ADF values within CNN-based and Swin transformer-based approaches, the proposed NSATP achieves better results (close to 1) compared to \textit{ArrivalNet}, which reflects the effect of non-stationarity recovery. 

\subsubsection{Ablation study}
In the proposed method, there are two compensation modules ($\tau_{\text{CNN}}$, $\mathbf{\Delta}_{\text{CNN}}$) in NSATP-1 (CNN) and four modules ($\tau^{1}_{\text{Swin}}$, $\mathbf{\Delta}^{1}_{\text{Swin}}$, $\tau^{2}_{\text{Swin}}$, $\mathbf{\Delta}^{2}_{\text{Swin}}$) in NSATP-2 (Swin). To investigate the contribution of each component, the ablation study is conducted and quantitative results are presented in Table~\ref{table_ablation}. Specifically, SS is the series stationarization. w/ and w/o mean with and without SS, respectively. Both represents that the model is equipped with $\tau^{1}_{\text{Swin}}$, $\mathbf{\Delta}^{1}_{\text{Swin}}$, $\tau^{2}_{\text{Swin}}$ and $\mathbf{\Delta}^{2}_{\text{Swin}}$. IN and OUT donate that  compensation modules ($\tau_{\text{CNN}}$, $\mathbf{\Delta}_{\text{CNN}}$) are inside each 2D block or only applied on the last 2D block of \textit{ArrivalNet}. In Table~\ref{table_ablation}, several findings are drawn. 

First, in comparison to the model with series stationarization, the model without it decreases the performance in all three metrics significantly. This phenomenon exists in both NSATP-1 (CNN) and NSATP-2 (Swin). It suggests that the non-stationary effect recovery approaches only work when the input is stationarized with the series stationarization. Second, comparing the results of \textit{ArrivalNet} and NSATP, only with the series stationarization is necessary for the sequential prediction but will lead to the over-stationarization issue. Third, for NSATP-1 (CNN), the compensation outside the 2D block is beneficial for performance. Fourth, for NSATP-2 (Swin), compared to only with $\tau^{1}_{\text{Swin}}$ and $\mathbf{\Delta}^{1}_{\text{Swin}}$, or only with $\tau^{2}_{\text{Swin}}$ and $\mathbf{\Delta}^{2}_{\text{Swin}}$, fully equipped with $\tau^{1}_{\text{Swin}}$, $\mathbf{\Delta}^{1}_{\text{Swin}}$, $\tau^{2}_{\text{Swin}}$ and $\mathbf{\Delta}^{2}_{\text{Swin}}$ achieve a further promotion.

\begin{table}[h!]
  \caption{The Quantitative Results of Ablation Experiments}
    \centering
    \label{table_ablation}
    \begin{tabular}{cccccc}
\toprule
 \textbf{Method}&  \textbf{SS}& -  &\textbf{RMSE} & \textbf{MAE} &\textbf{MAPE} \\ 
 \hline
\multirow{4}{*}{\parbox{2cm}{\centering NSATP-1 \\(CNN) }} 
& w/ & {IN}  & 56.2  & 36.8  & 2.485  \\ 
&w/&  {OUT}  & 55.8  & 36.4  & 2.47  \\ 
&w/o & {IN}  & 68.4  & 48.8  & 3.87  \\ 
& w/o&  {OUT}  & 67.3  & 47.3  & 3.54  \\
\hline
\multirow{6}{*}{\parbox{2cm}{\centering NSATP-2 \\(Swin) }} 
& w/ & Only $\tau^{1}_{\text{Swin}}$, $\mathbf{\Delta}^{1}_{\text{Swin}}$  & 55.7  & 38.1  & 2.62  \\ 
& w/&  Only $\tau^{2}_{\text{Swin}}$, $\mathbf{\Delta}^{2}_{\text{Swin}}$ & 56.4  & 38.3  & 2.67  \\ 
&w/ &  Both  & 54.9  & 37.5  & 2.60  \\ 
& w/o & Only $\tau^{1}_{\text{Swin}}$, $\mathbf{\Delta}^{1}_{\text{Swin}}$  & 66.4  & 46.1  & 3.48  \\ 
& w/o&  Only $\tau^{2}_{\text{Swin}}$, $\mathbf{\Delta}^{2}_{\text{Swin}}$ & 69.2  & 47.9  & 3.77  \\ 
&w/o &  Both  & 68.4  & 46.5  & 3.68  \\ 
 \bottomrule
\end{tabular}
\label{tab:merged_cells}
 \begin{tablenotes}
\item Note: SS: series stationarization, {IN}: compensation inside the 2D block, {OUT}: compensation outside the 2D block.
\end{tablenotes}
\end{table}

\section{Conclusion and future work}\label{conclusion}
To cope with the over-stationarization issue in the multi-step ATP, this work theoretically analyzes the causes of over-stationarity in the basic model \textit{ArrivalNet} and proposes two neural network-based non-stationarity effect recovery approaches. It guarantees the forecasting performance by the series stationarization and captures the instructive non-stationarity information from the raw sequence. The validation based on the large-scale city-wide dataset indicates that the proposed NSATP can promote the predictability of time series about the public transport ATP, which are naturally non-stationary. This work provides a new perspective for non-stationarity recovery in dealing with the over-stationarization issue.

This study mainly concentrates on the linearity of specific components in the basic model \textit{ArrivalNet} and the design of compensation modules, some assumptions may not apply to other models. Future works will focus on more general and model-agnostic solutions to the over-stationarization issue.

\appendix
\section{The Proof of Proposition~\ref{proposi_1}}\label{appendix_proposi_1}
In the series stationarization, the original time series $\mathbf{X} = [\mathbf{X}_1,\mathbf{X}_2,...,\mathbf{X}_{T}]^{\top}$ is converted into $\mathbf{X}^{\prime} = [\mathbf{X}^{\prime}_1,\mathbf{X}^{\prime}_2,...,\mathbf{X}^{\prime}_{T}]^{\top}$ based on mean $\pmb{\mu}\in \mathbb{R}^{C\times 1}$ and standard deviation $\pmb{\sigma}\in \mathbb{R}^{C\times 1}$, where $T$ is the length of input sequence and $C$ is the length of feature space. For each element $\mathbf{X}_i\in \mathbb{R}^{C\times 1}$ and $\mathbf{X}^{\prime}_i\in \mathbb{R}^{C\times 1}$, their relationship can be formulated as:
\begin{equation}
\mathbf{X}_i = \pmb{\sigma}\mathbf{X}^{\prime}_i + \pmb{\mu} \  \  \  i\in [1, T], t\in\mathbb{Z}
\end{equation}
where index $i$ donates the random element in $\mathbf{X}$ and $\mathbf{X}^{\prime}$. With Assumption~\ref{assum_1}, the vectorized standard deviation can be reduced as a scalar. Therefore, the following derivation is conducted,
\begin{equation}
\begin{aligned}
\mathbf{X}_i &= f\left(\sigma\mathbf{X}^{\prime}_i + \pmb{\mu}\right) \\
&= f(\sigma\mathbf{X}^{\prime}_i) + f(\pmb{\mu})\\
&=\sigma f(\mathbf{X}^{\prime}_i)+ f(\frac{1}{T}\sum_{j=1}^{T} \mathbf{X}_j) \\
&= \sigma f(\mathbf{X}^{\prime}_i) + \frac{1}{T}\sum_{j=1}^{T}f(\mathbf{X}_j)\\
&= \sigma f(\mathbf{X}^{\prime}_i) +\pmb{\mu}_{f(\mathbf{X})} \  \  \  i\in [1, T], t\in\mathbb{Z}
\end{aligned}
\end{equation}
where $\sigma$ is the scalar standard deviation and $\pmb{\mu}_{f(\mathbf{X})}$ is the mean of $f(\mathbf{X})$. Then,  $f(\mathbf{X})$ can be written as  $f(\mathbf{X})= \sigma f({\mathbf{X}}^{\prime})+\mathbf{1}\pmb{\mu}^{\top}_{f(\mathbf{X})}$ with $\mathbf{1}\in\mathbb{R}^{T\times 1}$.

\section{The proof of linearity in the fast Fourier transform}\label{appendix_FFT}
The FFT is an algorithm that computes the discrete fourier transform (DFT) of a sequence~\cite{brigham1988fast,sundararajan2001discrete}. To prove the linearity of FFT, we essentially need to demonstrate that the DFT itself is linear, since FFT is just a more efficient implementation of the DFT computation. The DFT of a sequence $\mathbf{X}$ with length $T$ is defined as:
\begin{equation}
    \mathbf{Y}[k] = \sum_{t=0}^{T-1} \mathbf{X}[n] e^{-i 2 \pi \frac{k t}{T}} \ \ \ \ k\in [0, T-1], k\in\mathbb{Z}
\end{equation} 
where $k$ and $t$ are the index of $k^{th}$ and $t^{th}$ element in the  transformed sequence $\mathbf{Y}$ and original input sequence $\mathbf{X}$. $i$ is the  imaginary unit. It transforms a sequence from the time domain into the frequency domain, encoding both amplitude and phase information at discrete frequency components. The linearity of the DFT is based on two properties: additivity and homogeneity, which are detailed in Property~\ref{property_1}. To prove additivity, we consider two sequences $\mathbf{X}_1$ and $\mathbf{X}_2$ both with length $T$, and show that the DFT of their sum is the sum of their DFTs. 
\begin{equation}\label{fft_additivity}
\begin{aligned}
&\text{DFT}\{\mathbf{X}_1 + \mathbf{X}_2\}[k] \\
&= \sum_{t=0}^{T-1} (\mathbf{X}_1[t] + \mathbf{X}_2[t]) e^{-i 2 \pi \frac{k t}{T}} \\
&= \sum_{t=0}^{T-1} \mathbf{X}_1[t] e^{-i 2 \pi \frac{k t}{T}} + \sum_{t=0}^{T-1} \mathbf{X}_2[t] e^{-i 2 \pi \frac{k t}{T}} \\
&= \text{DFT}\{\mathbf{X}_1\}[k] + \text{DFT}\{\mathbf{X}_2\}[k] \ \ \ \ k\in [0, T-1], k\in\mathbb{Z}
\end{aligned}
\end{equation}

Similarly, to prove homogeneity, we consider a scalar $a$ and a sequence $\mathbf{X}$ and the scaled sequence's DFT is:
\begin{equation}\label{fft_homogeneity}
\begin{aligned}
\text{DFT}\{a \mathbf{X}\}[k] &= \sum_{n=0}^{T-1} (a \mathbf{X}[t]) e^{-i 2 \pi \frac{k t}{T}} \\
&= a \sum_{t=0}^{T-1} \mathbf{X}[t] e^{-i 2 \pi \frac{k t}{T}} \\
&= a \text{DFT}\{\mathbf{X}\}[k] \ \ \ \ k\in [0, T-1], k\in\mathbb{Z}
\end{aligned}
\end{equation}

With the additivity and homogeneity in Eq.~\eqref{fft_additivity} and Eq.~\eqref{fft_homogeneity}, the linearity of FFT is proved. Based on Proposition~\ref{proposi_1}, the FFT of normalized $\mathbf{X}^{\prime}$ is expressed as follows:
\begin{equation}
    f_{\text{FFT}}({\mathbf{X}}^{\prime})= \frac{1}{\sigma}(f_{\text{FFT}}(\mathbf{X})-\mathbf{1}\pmb{\mu}^{\top}_{f_{\text{FFT}}(\mathbf{X})})
\end{equation}
where $\mathbf{X}$ is the original sequence. $\pmb{\mu}_{f_{\text{FFT}}(\mathbf{X})}=\frac{1}{T}\sum_{i=1}^{T}f(\mathbf{X}_i)$.

\section{Augmented Dickey-Fuller Test}\label{appendix_adf}
The Augmented Dickey-Fuller test is a common statistical test used to determine whether a given time series is stationary~\cite{elliott1992efficient}. This test is particularly useful in identifying the presence of a unit root in a series, which indicates non-stationarity. With the input time series $\mathbf{y} = [y_1, y_2,..., y_n]^{\top}$, the general description of ADF regression is as follows:
\begin{equation}
\Delta y_t = \alpha + \beta t + \gamma y_{t-1} + \sum_{i=1}^{p-1} \delta_i \Delta y_{t-i} + \epsilon_t
\end{equation}
where $\Delta y_t = y_t - y_{t-1}$ is the first difference of $y_t$. $\alpha$  is a constant of intercept. $\beta t$ represents the time trend. $\gamma$ is the coefficient on the lagged value of the series, which is typically expected to be less than zero in the presence of stationarity. $\delta_i$ are the coefficients for the lagged differences of the series and $\epsilon_t$ is the error term. $p$  is the number of lagged first differences included in the regression, determined by various criteria like Akaike Information Criterion (AIC)~\cite{akaike2011akaike}. The ADF statistic is derived from the coefficient $\gamma$, which is estimated by ordinary least squares (OLS) method~\cite{craven2011ordinary}. The test statistic is then typically computed as $\frac{\hat{\gamma}}{\text{SE}(\hat{\gamma})}$, where $\text{SE}(\cdot)$ is the standard error of the estimated coefficient.

\bibliographystyle{IEEEtran}
\bibliography{ref.bib}
\end{document}